\documentclass[]{afu}
\usepackage{xcolor}
\usepackage{booktabs}
\usepackage{multirow}
\usepackage{xcolor}
\usepackage{colortbl}
\usepackage{graphicx}

\usepackage{amssymb}
\usepackage{amsmath}
\usepackage{array}
\usepackage{tabularx}
\usepackage{makecell}
\usepackage{algorithm}
\usepackage{algpseudocode}
\usepackage[most]{tcolorbox}
\usepackage{listings}
\usepackage{fancyvrb}
\usepackage{enumitem}
\usepackage{caption}
\usepackage{subcaption}
\usepackage{url}
\usepackage{hyperref}
\usepackage{mathtools}
\usepackage{amsthm}
\usepackage{tcolorbox} 
\usepackage{enumitem}
\usepackage{bbm}
\usepackage[table]{xcolor}
\usepackage[dvipsnames]{xcolor}
\usepackage{booktabs}
\usepackage{makecell}
\usepackage[table]{xcolor}
\usepackage{array}

\usepackage{dsfont}

\newcounter{prompt}
\renewcommand{\theprompt}{\arabic{prompt}}
\usepackage{pifont}
\usepackage{graphicx}
\usepackage{longtable}
\usepackage{listings}
\usepackage[numbers]{natbib}
\definecolor{codegreen}{rgb}{0,0.6,0}
\definecolor{codegray}{rgb}{0.5,0.5,0.5}
\definecolor{codepurple}{rgb}{0.58,0,0.82}
\definecolor{backcolour}{rgb}{0.95,0.95,0.92}
\definecolor{promptcolor}{HTML}{D1D0F2}
\definecolor{promptcolorheader}{HTML}{bdbcec}

\definecolor{GainCell}{HTML}{2E7D32}
\definecolor{DropCell}{HTML}{C62828}
\newcommand{\poscell}[2]{\cellcolor{GainCell!#1}#2}
\newcommand{\negcell}[2]{\cellcolor{DropCell!#1}#2}

\newcounter{dataexample}
\renewcommand{\thedataexample}{\arabic{dataexample}}

\newcommand{\promptbox}[2]{
\begin{tcolorbox}[
top=0.3em,bottom=0.3em,left=0.5em,right=0.5em,
toptitle=0.3em,bottomtitle=0.2em,boxsep=0pt,
colframe=promptcolorheader,colback=promptcolor!50,boxrule=0.5pt,
]
\footnotesize
\end{tcolorbox}
}
\lstdefinestyle{mystyle}{
    backgroundcolor=\color{backcolour},   
    commentstyle=\color{codegreen},
    keywordstyle=\color{magenta},
    numberstyle=\tiny\color{codegray},
    stringstyle=\color{codepurple},
    basicstyle=\ttfamily\footnotesize,
    breakatwhitespace=false,         
    breaklines=true,                 
    captionpos=b,                    
    keepspaces=true,                 
    numbers=left,                    
    numbersep=5pt,                  
    showspaces=false,                
    showstringspaces=false,
    showtabs=false,                  
    tabsize=2
}

\title{MMVistaReason: Toward Open-Data and Post-Training Recipes for Multimodal Reasoning}

\author[1,2]{Juekai Lin$^*$}
\author[2,3]{Honglin Lin$^*$}
\author[1]{Yuqian Yuan}
\author[4]{Xiaolong Wu}
\author[1]{Jie Cao}
\author[1]{Liang Liang}
\author[1]{Yunqi Cao}
\author[2]{Yun Zhu}
\author[1]{Wenqiao Zhang}
\author[2]{Lijun Wu}

\affiliation[1]{Zhejiang University}
\affiliation[2]{Ant Group, AFu Team}
\affiliation[3]{Shanghai Jiao Tong University}
\affiliation[4]{Tongji University}

\abstract{
Open multimodal reasoning models have benefited from large-scale reasoning supervision, yet reliable post-training remains challenging, with bottlenecks extending beyond model scaling to uneven data quality, inefficient supervision construction, imbalanced difficulty, and cross-domain interference. We introduce \textbf{MMVistaReason (MVR)}, an open-data and
post-training recipe that addresses these challenges from three perspectives:
(1) \textbf{broader capability coverage}, which organizes multimodal reasoning into two complementary groups---Analytical reasoning, emphasizing structured and knowledge-intensive problem solving, and Real-World reasoning, emphasizing visual perception and spatial grounding;
(2) \textbf{efficient SFT and RL data construction}, which standardizes heterogeneous open data through staged cleaning and annotation, combines difficulty-aware cascaded teacher distillation with answer-likelihood-based trajectory selection to construct MVR-SFT-528K, and applies scale-specific frontier filtering for MVR-RL-63K; and
(3) \textbf{specialize-then-integrate training}, which trains complementary RL experts for each reasoning group and consolidates their capabilities through multi-teacher on-policy distillation (MOPD).
Our analyses reveal a capacity-dependent interaction between supervision difficulty, trajectory quality, and model capacity, showing that smaller students benefit more from selected supervision while larger students are more robust to trajectory variation and mixed-domain interference. We find that mixed-domain RL introduces benchmark-level negative transfer, whereas MOPD provides more consistent capability integration, with the preferred KL direction varying across model scales. Across 15 multimodal benchmarks, MVR-4B achieves an average score of 72.8, outperforming Qwen3.5-9B (Instruct) and MMFineReason-8B while using about 70\% fewer samples than MMFineReason. Scaling to 9B further improves the average to 74.4, surpassing Qwen3.5-35B-A3B (Instruct). Overall, MMVistaReason demonstrates that systematic open-data construction and capacity-aware post-training provide a practical and scalable path toward reliable multimodal reasoning models.
}

\date{\today}
\metadata[Equal Contributions]{Juekai Lin, Honglin Lin}
\correspondence{Lijun Wu, \email{wlj554215@antgroup.com}}

\metadata[Project]{\url{https://github.com/JackieForest/MMVistaReason}}

\begin{document}

\maketitle

\section{Introduction}

Open multimodal reasoning has rapidly progressed from visual instruction tuning to large-scale reasoning supervision. Recent long-form reasoning datasets have substantially improved mathematical and scientific reasoning in open vision-language models (VLMs) \citep{leng2025mmr1,lin2026mmfinereason,zhang2026openmmreasoner,lu2024mathvista,tao2026omniscience,lin2026scientific}, as well as diagrammatic and spatial reasoning \citep{xu2025llava,bansal2026honeybee,liu2026spatial,yuan2026lmms}. In particular, MMFineReason (MFR) shows that carefully curated supervision enables compact VLMs to compete with much larger models, while difficulty-aware subsets can approach full-data performance. Meanwhile, advances in large language models (LLMs) have highlighted systematic post-training recipes, including high-quality instruction data \citep{guha2026openthoughts,ye2025limo}, rejection sampling \citep{koh2026adastar,khairi2026making}, verifiable feedback \citep{chen2026longrlvr}, and reinforcement learning \citep{wen2026reinforcement}. Extending these principles to multimodal reasoning remains challenging because open multimodal data vary in visual context, quality, answerability, format, and difficulty. Therefore, reliable multimodal reasoning requires not only larger models or more data, but systematic recipes for constructing supervision, adapting it to model capacity, and integrating complementary reasoning capabilities.

Despite recent progress in multimodal reasoning data construction and post-training
\citep{leng2025mmr1,lin2026mmfinereason,zhang2026openmmreasoner,xu2025llava,bansal2026honeybee},
three key challenges remain:

\begin{itemize}

\item \textbf{Broader and more balanced capability coverage.}
Existing multimodal reasoning data are still dominated by analytical tasks such as mathematics, science, and structured visual reasoning, while real-world capabilities including perception, counting, spatial understanding, and GUI grounding receive comparatively less emphasis. A reliable multimodal reasoner should therefore develop strong and balanced capabilities across both analytical and real-world scenarios.

\item \textbf{Data efficiency and adaptive distillation.}
Multimodal samples vary in difficulty and trajectory quality, while conventional pipelines often distill supervision from a single large teacher at high inference cost. Moreover, SFT and RL require different supervision: SFT benefits from reliable trajectories matched to student capacity, whereas RL favors frontier prompts that remain learnable under the current policy, motivating efficient and stage-specific data construction.

\item \textbf{Cross-domain interference and capability integration.}
Training heterogeneous reasoning domains under a single mixed objective can introduce negative transfer across tasks \citep{sachdeva2026tasktransfer}. While specialist models can better capture domain-specific capabilities, integrating these complementary strengths into a unified generalist remains underexplored. On-policy distillation methods \citep{agarwal2024policy,ma2026mopd} provide a promising direction for capability integration on student-generated trajectories.

\end{itemize}

To address these challenges, we introduce \textbf{MMVistaReason (MVR)}, a systematic framework that advances existing multimodal reasoning recipes in both data construction and model optimization. We build a broader and more balanced corpus spanning two complementary groups: \textbf{Analytical}, covering mathematics, science, visual logic, and charts with an emphasis on structured, knowledge-intensive, and symbolic reasoning; \textbf{Real-World}, covering general perception, counting, spatial understanding, and GUI grounding with greater emphasis on fine-grained visual perception and spatial grounding.

\textbf{For SFT}, we develop a difficulty-aware cascaded distillation pipeline that identifies challenging samples and routes unresolved examples to progressively stronger same-family teachers. Verified correct trajectories are selected through Answer-Likelihood Best-of-N, yielding reliable and difficulty-stratified supervision. This enables us to study how supervision difficulty and trajectory quality interact with model capacity.
\textbf{For RL}, we use the rollout pass rates of the SFT model at each scale to identify frontier prompts with learning potential. We balance these prompts across the Analytical and Real-World groups and train complementary RL experts specialized for each reasoning group.
\textbf{For MOPD}, we consolidate the capabilities of both experts into a unified generalist through multi-teacher on-policy distillation, where routed experts provide token-level supervision on student-induced trajectories. We investigate both forward- and reverse-KL objectives for capability integration.

Extensive experiments demonstrate the effectiveness of MMVistaReason from three perspectives. 
(1) \textbf{Overall performance:} Across 15 multimodal benchmarks, MVR-4B achieves an average of 72.8, outperforming Qwen3.5-9B (Instruct) and MFR-8B while using half the model size and approximately 70\% fewer SFT samples than MFR. Scaling to 9B improves the average to 74.4, surpassing Qwen3.5-35B-A3B (Instruct).
(2) \textbf{Data construction:} Our analyses reveal a capacity-dependent interaction between supervision difficulty, data scale, and trajectory quality. Smaller students benefit more from broader selected supervision, whereas the 9B student matches full-data performance using only a 124K moderately difficult subset. Answer-Likelihood Best-of-N provides more reliable trajectory selection than random or length-based strategies across model scales.
(3) \textbf{Training strategy:} Complementary RL experts develop domain-specific strengths, while MOPD consolidates these capabilities into unified models, improving the SFT generalists from 70.8 to 72.8 at 4B and from 72.3 to 74.4 at 9B. Compared with mixed-domain RL, MOPD provides consistent improvements with reduced cross-domain interference, while the preferred KL direction varies with model scale. Together, these results validate the efficient data construction and specialize-then-integrate post-training recipes of MMVistaReason.

Our contributions are summarized as follows:

\begin{itemize}

\item We introduce \textbf{MMVistaReason (MVR)}, a systematic recipe suite that broadens multimodal reasoning from a narrower task focus to balanced \textbf{Analytical} and \textbf{Real-World} scenarios, with MVR-4B outperforming MFR-8B using half the model size and about 70\% fewer SFT samples.

\item We develop an efficient SFT and RL data construction pipeline combining cascaded teacher distillation, Answer-Likelihood Best-of-N selection, and scale-specific frontier filtering to build two compact, high-quality resources: \textbf{MVR-SFT-528K} and \textbf{MVR-RL-63K}. Our analysis further reveals a capacity-dependent relationship between supervision difficulty and data scale.

\item We develop a \textbf{specialize-then-integrate} training recipe with complementary RL experts and multi-teacher on-policy distillation (MOPD), showing that it reduces mixed-domain interference and provides consistent capability integration across scales, with scale-dependent preferences between forward- and reverse-KL objectives.

\end{itemize}

\begin{figure*}[t]
    \centering
    \includegraphics[width=\textwidth]{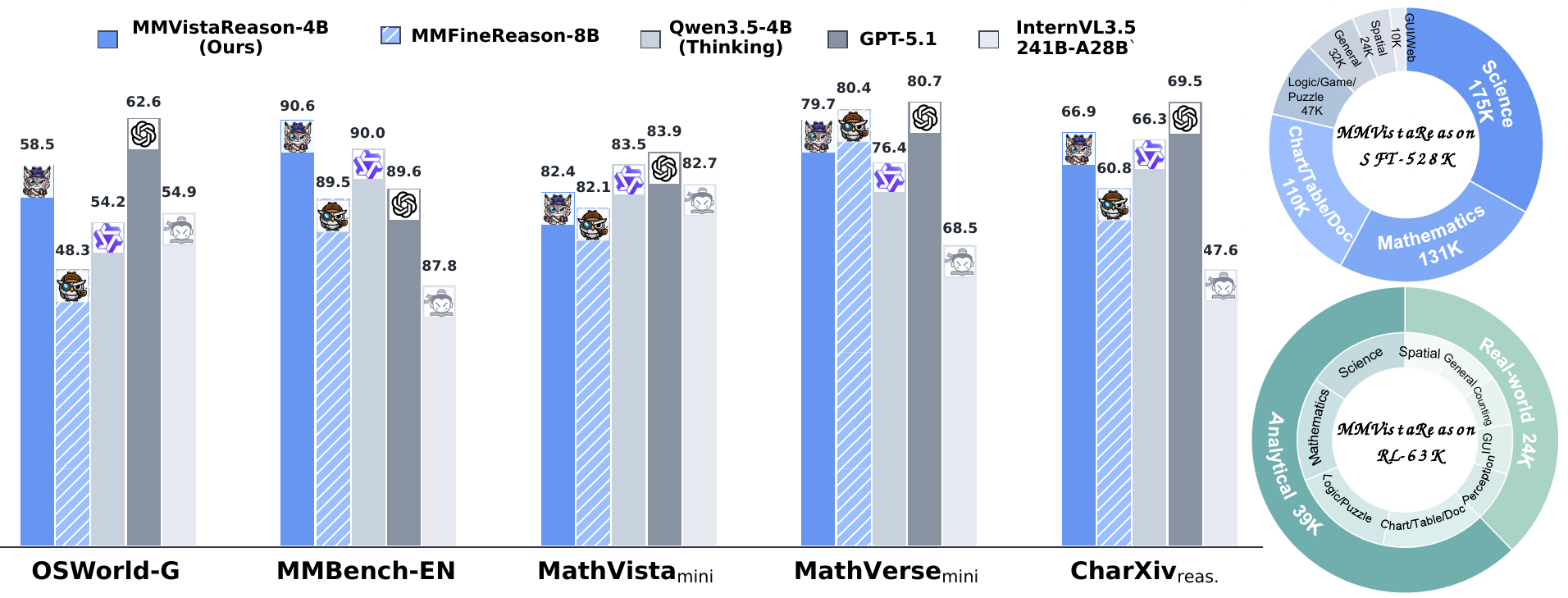}
    \caption{
    Overview of MMVistaReason benchmark performance.
    Left: MMVistaReason achieves strong scores against representative baselines; right: MMVistaReason-SFT-528K and MMVistaReason-RL-63K cover diverse reasoning domains and complementary training groups.
    }
    \label{fig:mfr_pro_overview}
\end{figure*}

\section{Related Work}

\subsection{Multimodal Reasoning Recipes}

Early multimodal training recipes primarily relied on LLM-generated visual instructions and supervised fine-tuning over broad task mixtures, establishing general-purpose visual instruction-following capabilities
\citep{liu2023visual,li2024llava}.
Subsequent studies introduced domain-specific or rationale-enriched supervision, including MathV360K, LLaVA-CoT-100K, and MAmmoTH-VL, to provide more explicit reasoning signals and improve performance in specialized scenarios
\citep{shi2024math,xu2025llava,guo2025mammoth}.
More recent data-centric efforts further investigate the effects of context sources, data composition, reasoning quality, and difficulty-aware sample selection, shifting attention from data scaling toward deliberate curation of multimodal reasoning supervision
\citep{bansal2026honeybee,lin2026mmfinereason}. Beyond supervised fine-tuning, recent works have explored reinforcement learning with outcome-level or step-level rewards for multimodal reasoning
\citep{huang2026vision,mmk12,zhang2025r1}.
R1-Onevision and OpenMMReasoner further adopt two-stage SFT-RL pipelines, combining supervised initialization with subsequent policy optimization to improve multimodal reasoning performance
\citep{leng2025mmr1,yang2025r1,zhang2026openmmreasoner}.
However, existing open multimodal reasoning recipes typically optimize a single generalist model over aggregated mixed-domain data, without explicitly addressing the heterogeneous capabilities required by different reasoning scenarios
\citep{lin2026mmfinereason,yang2025r1,zhang2026openmmreasoner}.
Meanwhile, controlled studies have shown that VLM fine-tuning can exhibit both positive and negative cross-task transfer, suggesting limitations of uniform joint optimization and motivating more adaptive training strategies
\citep{sachdeva2026tasktransfer}.
Our work builds upon these efforts by combining systematic data construction, complementary expert specialization, and capability integration for reliable multimodal reasoning post-training.

\subsection{Knowledge Distillation}

Knowledge distillation transfers capabilities from a strong teacher to a compact student, traditionally by matching softened output distributions
\citep{hinton2015distilling, xu2024survey}.
For autoregressive models, sequence-level distillation further extends this paradigm by training students on complete teacher-generated sequences rather than independent token labels
\citep{kim2016sequence}.
Recent LLM distillation methods explore objectives tailored for generative models, such as MiniLLM's reverse-KL optimization
\citep{gu2024minillm}.
In multimodal reasoning, offline distillation commonly leverages strong teachers to generate structured rationales or reasoning trajectories, which are then used as fixed supervised fine-tuning targets
\citep{xu2025llava,guo2025mammoth,hsieh2023distilling}.
However, in practical settings, this paradigm introduces a mismatch between teacher-generated trajectories and student-induced states during autoregressive generation
\citep{agarwal2024policy}.

On-policy distillation addresses this issue by sampling trajectories from the student policy and querying teachers for token-level supervision on student-induced prefixes
\citep{agarwal2024policy}.
Recent studies further extend this paradigm through self-distillation, efficient prefix-level supervision, and relaxed imitation objectives
\citep{zhao2026self,zhang2026fast,ko2026scaling,yuan2026vision}.
Beyond single-teacher transfer, MOPD integrates multiple domain-specific RL teachers into a unified student
\citep{ma2026mopd,gao2026open,yang2026nemotron,huang2026kat,team2026mach}.
On-policy distillation has also been explored in multimodal models: VOLD combines GRPO with distillation from a text-only reasoning teacher, Visual-Advantage OPD emphasizes tokens containing fine-grained visual evidence, and OPOD coordinates specialized teachers across text, image, and audio modalities
\citep{bousselham2026vold,liu2026visual,zhao2026opod}.
Building upon these advances, our work employs multi-teacher on-policy distillation to consolidate complementary real-world and analytical RL experts into unified multimodal generalists.

\section{MMVistaReason Recipes}

In this section, we present the MMVistaReason recipe framework, which consists of three components: Data Cleaning Recipe, Data Construction Recipe, Model Training and Integration Recipe. Together, they provide a unified pipeline from open-data curation to SFT and RL data construction, domain-specific RL training, and multi-teacher OPD integration.

\begin{figure*}[t]
    \centering
    \includegraphics[width=\textwidth]{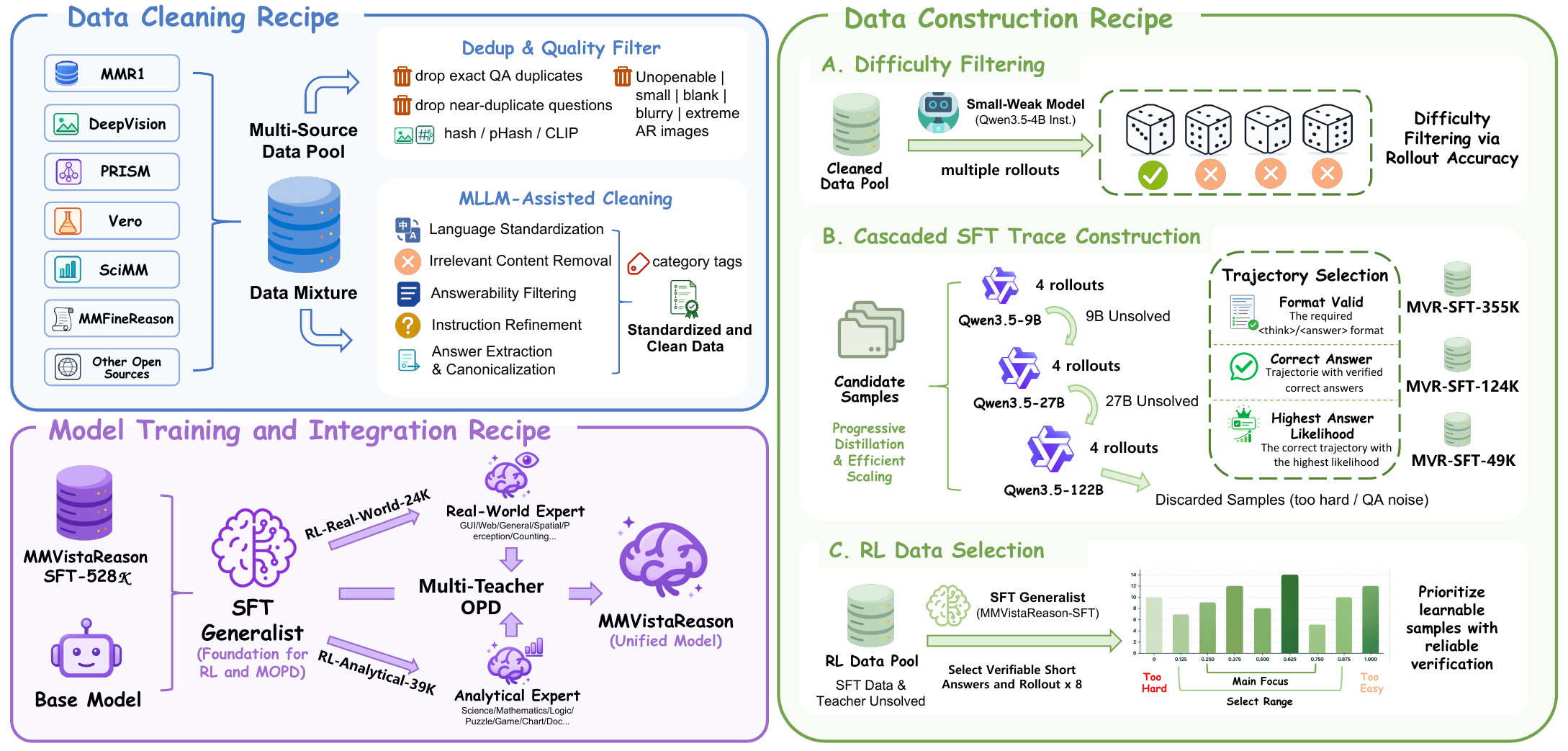}
    \caption{
Overview of the MMVistaReason Recipes.
It introduces a unified pipeline covering data cleaning, reasoning trace construction, reinforcement learning, and multi-teacher integration, enabling scalable training of reliable multimodal reasoning models from open data.
}
    \label{fig:mfr_pro_overview}
\end{figure*}

\subsection{Data Cleaning Recipe}

We aggregate diverse open-source multimodal VQA-style datasets, which contain substantial redundancy and noise due to their heterogeneous origins. We therefore adopt a staged cleaning pipeline combining rule-based filtering with MLLM-assisted semantic cleaning and annotation.

\paragraph{Deduplication and Image Quality Filtering.}
We first remove redundant and visually invalid samples. For deduplication, samples with the same image and identical question-answer pairs are removed, while those sharing the same image with highly similar questions are treated as near-duplicates; different questions associated with the same image are retained. We combine image SHA-256 hashes, perceptual hashes (pHash), and CLIP-based semantic similarity for duplicate detection. We further discard corrupted, low-resolution, blank, blurry, or extreme-aspect-ratio images.

\paragraph{MLLM-Assisted Cleaning and Annotation.}
We then employ Qwen3.5-122B-A10B, a strong multimodal model, to perform semantic-level cleaning and normalization through six steps:
\begin{itemize}
\item \textbf{Language Standardization.} Translate non-English questions into English while preserving formulas, variables, option labels, units, and visual identifiers.

\item \textbf{Irrelevant Content Removal.} Remove unrelated links, advertisements, watermarks, problem indices, score annotations, and other distracting artifacts.

\item \textbf{Answerability Filtering.} Remove samples that cannot be reliably answered from multimodal input, such as incomplete questions, external action requirements, or non-answering tasks.

\item \textbf{Instruction Refinement.} Rewrite low-quality meta-instructions that unnecessarily discourage reasoning, while preserving the original task semantics.

\item \textbf{Answer Extraction.}
Extract the final answer from noisy answer fields containing text or intermediate reasoning, providing clean targets for subsequent normalization and verification.

\item \textbf{Answer Canonicalization.} Normalize answers into consistent and verifiable forms, particularly by recovering option labels for multiple-choice questions whose answers contain option content.

\end{itemize}

\paragraph{Data Annotation and Standardization.}
After cleaning, we annotate each sample with its reasoning domain and question type, providing structured labels for subsequent domain-aware data organization and sampling. We then convert all samples into a unified schema containing normalized image-question-answer fields, annotation labels, and source metadata (representative data examples are provided in Appendix~\ref{app:data_examples}). This standardized representation enables consistent difficulty filtering, reasoning-trace construction, balanced sampling, and downstream SFT and RL training.

\subsection{Data Construction Recipe}

\begin{figure*}[t]
    \centering
    \includegraphics[width=\textwidth]{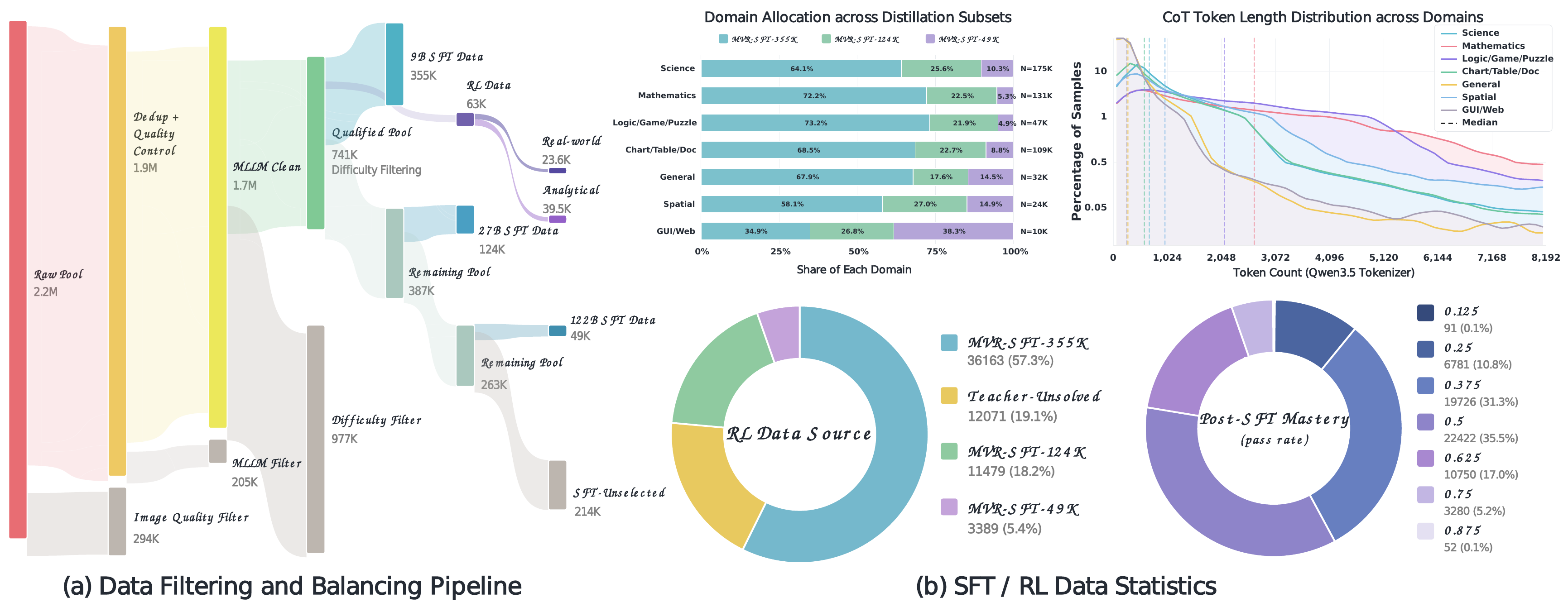}
   \caption{
Data flow and statistics of MMVistaReason.
(a) Data filtering and balancing from the raw pool to SFT and RL data.
(b) Statistics of MMVistaReason datasets.
}
    \label{fig:data_statistics}
\end{figure*}

\begin{table*}[t]
  \centering
  \fontsize{6.8pt}{7.8pt}\selectfont
  \begin{tabular}{llrrrrrrrrrr}
  \toprule
  \textbf{Dataset} & \textbf{Distill-model} & \textbf{\#Samples} & \textbf{Tokens} & \textbf{Mean} & \textbf{Std.} & \textbf{Median} &
  \textbf{P25} & \textbf{P75} & \textbf{P95} & \textbf{Min} & \textbf{Max} \\
  \midrule
  MVR-SFT-355K & Qwen3.5-9B   & 354,886 & 631.72M  & 1,780.06 & 2,203.43 & 900   & 436 & 2,195 & 6,316  & 26 & 16,374 \\
  MVR-SFT-124K & Qwen3.5-27B  & 124,010 & 309.30M  & 2,494.13 & 3,290.11 & 1,031 & 440 & 2,998 & 10,414 & 45 & 16,375 \\
  MVR-SFT-49K  & Qwen3.5-122B & 48,913  & 103.04M  & 2,106.55 & 3,120.21 & 670   & 340 & 2,313 & 9,657  & 39 & 16,381 \\
  \midrule
  \textbf{MVR-SFT-528K} & \textbf{All} & \textbf{527,809} & \textbf{1.04B} & \textbf{1,978.09} & \textbf{2,607.77} & \textbf{906} &
  \textbf{426} & \textbf{2,359} & \textbf{7,625} & \textbf{26} & \textbf{16,381} \\
  \bottomrule
  \end{tabular}
  \caption{Statistics of MMVistaReason (MVR)-SFT-528K and its sub-datasets. Each sub-dataset is distilled from the teacher model that first
  solves the sample in the cascaded rejection sampling pipeline. Token statistics are computed over model responses using the Qwen3.5 tokenizer.}
  \label{tab:sft_data_stats}
\end{table*}

Based on the cleaned data pool, we construct \textbf{MMVistaReason-SFT-528K} and \textbf{MMVistaReason-RL-63K} for subsequent model training, with the latter comprising scale-specific RL subsets for 4B and 9B models. For trajectory verification, all generated responses are judged by \textbf{Qwen3.5-27B}, prioritizing answer correctness while also considering overall reasoning consistency.

\textbf{SFT Data Construction.}
We construct SFT trajectories with a \textbf{difficulty-aware cascaded pipeline}, first filtering trivial samples and assigning harder samples to progressively stronger teachers. This design focuses supervision on challenging examples while reducing unnecessary calls to large teacher models.

We first use Qwen3.5-4B (Instruct) for \textbf{model-aware difficulty filtering}. For each sample $z=(x,a)$, we generate four responses and compute the verified pass rate:
\[
p_{\rm 4B}(z)=\frac{1}{4}\sum_{k=1}^{4}V(y_k,a),
\]
where $V(\cdot)$ denotes the verification function. We retain only samples with $p_{\rm 4B}(z)=0$, i.e., those that fail all four rollouts, thereby filtering out trivial samples before reasoning-trace construction.

For the retained hard samples, we apply \textbf{cascaded rejection sampling} with progressively stronger same-family teachers: Qwen3.5-9B, 27B, and 122B (instruct mode). For each sample $z=(x,a)$, teacher $T_j$ generates $K=4$ responses, and the verified correct set is defined as
\[
\mathcal{Y}_j(z)=\{y_{j,k}\mid V(y_{j,k},a)=1,\; k=1,\ldots,K\}.
\]
The sample is assigned to the first teacher that produces at least one correct response:
\[
j^*(z)=\min\{j\mid |\mathcal{Y}_j(z)|>0\}.
\]
Samples that remain unsolved by all teachers are discarded, as they may be overly difficult, ambiguous, or noisy. For the selected teacher, we further perform \textbf{Answer-Likelihood Best-of-N} trajectory selection. We first retain trajectories with verified correct answers and valid response formats, requiring complete \texttt{<think>...</think>} reasoning and \texttt{<answer>...</answer>} final-answer tags. If only one valid trajectory remains, it is directly selected. When multiple valid trajectories are available, we rank them by the normalized log-likelihood of the final answer tokens under the corresponding teacher:
\[
y^*(z)=
\arg\max_{y\in\widetilde{\mathcal{Y}}_{j^*(z)}}
\frac{1}{|a|}
\sum_{t=1}^{|a|}
\log p_{T_{j^*(z)}}(a_t\mid x,r_y,a_{<t}),
\]
where $\widetilde{\mathcal{Y}}_{j^*(z)}$ denotes the set of correct and format-valid trajectories, and $r_y$ denotes the reasoning trace preceding the final answer. This normalization reduces the bias introduced by answer length and favors trajectories that more confidently support the verified final answer. 

The selected reasoning trajectories constitute \textbf{MMVistaReason-SFT-528K}, comprising three subsets constructed by Qwen3.5-9B, 27B, and 122B through the cascaded pipeline with reduced inference cost.

\textbf{RL Data Construction.}
For RL data construction, we start from the hard pool retained by the 4B difficulty filter, ensuring non-trivial candidates. We remove samples with overly long or difficult-to-verify answers for reliable correctness evaluation. We then use the corresponding SFT-trained model $\pi_{\rm SFT}^{(s)}$ at each scale $s$ to identify frontier samples that the current policy solves inconsistently. For each answer-verifiable sample $z=(x,a)$, we generate eight rollouts and compute the verified pass rate:
\[
p_{\rm SFT}^{(s)}(z)=\frac{1}{8}\sum_{i=1}^{8}V(y_i,a),
\quad
y_i\sim\pi_{\rm SFT}^{(s)}(\cdot\mid x),
\]
where $V(\cdot)$ denotes the verification function. We retain samples within scale-specific frontier interval:
\[
\mathcal{D}_{\rm frontier}^{(s)}
=
\left\{
z\mid
\tau_{\rm low}\leq p_{\rm SFT}^{(s)}(z)\leq\tau_{\rm high}
\right\}.
\]
In practice, with eight rollouts, we retain samples solved between one and seven times, while prioritizing those solved two to six times. This removes fully failed samples, which are often overly difficult or noisy, and fully solved samples, which provide limited additional learning signals.

According to the reasoning characteristics of tasks, we perform balanced sampling across reasoning domains at each scale and organize the selected frontier samples into two complementary groups:
\[
\mathcal{D}_{\rm RL}^{(s)}
=
\mathcal{D}_{\text{real-world}}^{(s)}
\cup
\mathcal{D}_{\rm analytical}^{(s)}.
\]
Here, $\mathcal{D}_{\text{real-world}}^{(s)}$ covers general and spatial visual reasoning, including perception, counting, GUI-related tasks, and real-world visual scenarios, while $\mathcal{D}_{\rm analytical}^{(s)}$ includes mathematics, science, logic/puzzle, and chart/document reasoning. The resulting scale-specific subsets form \textbf{MMVistaReason-RL-63K}, which is used to train the complementary RL experts and multi-teacher OPD.

\textbf{Data Analysis.}
Figure~\ref{fig:data_statistics} summarizes the overall data flow and statistics of MMVistaReason.

For the SFT corpus, different teacher-derived subsets and reasoning domains show clear diversity. The three subsets are not simply scaled replicas: MVR-SFT-355K dominates most reasoning-intensive domains, contributing 64.1–73.2\% of Science, Mathematics, Logic/Game/Puzzle, and Chart/Table/Doc, whereas MVR-SFT-49K places greater emphasis on GUI/Web (38.3\%) and MVR-SFT-124K contributes relatively more to Spatial data (27.0\%). This complementary allocation indicates that cascaded teacher distillation provides supervision with varying domain coverage and reasoning complexity beyond data scale. At the domain level, Science, Mathematics, and Chart/Table/Doc together account for 78.61\% of the corpus. Mathematics and Logic/Game/Puzzle generally contain longer reasoning traces, while General and GUI/Web are much shorter. Overall, MMVistaReason-SFT covers diverse reasoning formats, ranging from short-form visual understanding to long-form analytical reasoning.

For the RL data, we analyze its source composition and Post-SFT Mastery distribution across the scale-specific subsets.
MVR-RL-63K is mainly drawn from the three SFT-derived subsets, which together contribute 80.9\%, while Teacher-Unsolved accounts for 19.1\%.
This indicates that RL training is not simply a continuation over SFT data, but also introduces frontier samples that were not converted into verified SFT trajectories by teacher distillation.
We define Post-SFT Mastery as the pass rate of the corresponding SFT model over eight rollouts.
Most RL samples fall into the middle mastery range of 0.375--0.625, accounting for 83.83\% of the RL set.
These samples are neither trivial nor nearly unsolvable; instead, they are cases where the corresponding SFT model can sometimes answer correctly but has not mastered the reasoning process stably.
Therefore, MVR-RL focuses on policy-relative frontier examples suitable for further expert specialization through reinforcement learning.

Overall, the SFT corpus provides broad and diverse reasoning supervision, while the RL corpus focuses on post-SFT frontier samples with remaining improvement potential.
This complementary design supports our staged training recipe: SFT first builds general multimodal reasoning ability, and RL further strengthens unstable reasoning behaviors.

\subsection{Model Training and Integration Recipe}

\textbf{Full-data SFT.}
We first train unified reasoning models on the full \textbf{MMVistaReason-SFT-528K} corpus. Given an input $x$ and its selected reasoning trajectory $y$, the SFT objective is
\[
\mathcal{L}_{\rm SFT}
=
-\mathbb{E}_{(x,y)\sim\mathcal{D}_{\rm SFT}}
\log \pi_\theta(y\mid x).
\]
This stage establishes a general multimodal reasoning policy that serves as the initialization for subsequent expert specialization and capability integration.

\textbf{Expert Specialization via RL.}
Starting from the corresponding SFT model, we train two complementary RL experts on scale-specific subsets of \textbf{MMVistaReason-RL-63K}, corresponding to \textbf{Real-World} Visual Reasoning and \textbf{Analytical} Visual Reasoning. For each model scale $s$ and reasoning group $g\in\{\text{real-world},\mathrm{analytical}\}$, we initialize the expert from $\pi_{\rm SFT}^{(s)}$ and optimize it on the subset:
\[
\pi_{\phi_g}^{(s)}
\leftarrow
\mathrm{RL}
\left(
\pi_{\rm SFT}^{(s)},
\mathcal{D}_{\rm RL}^{(s,g)}
\right).
\]

We adopt \textbf{Group Sequence Policy Optimization (GSPO)}~\cite{gspo}, which performs policy optimization with sequence-level importance ratios to reduce the token-level importance bias in GRPO and improve overall training stability. The optimization objective is defined as:
\[
\mathcal{J}_{\rm GSPO}(\phi_g)
=
\mathbb{E}
\left[
\frac{1}{G}
\sum_{i=1}^{G}
\min
\left(
s_i(\phi_g)\widehat{A}_{i},
\mathrm{clip}\!\left(s_i(\phi_g),1-\epsilon,1+\epsilon\right)
\widehat{A}_{i}
\right)
\right],
\]
where $s_i(\phi_g)$ denotes the importance ratio computed from the sequence likelihood, $\widehat{A}_{i}$ is the group-relative advantage estimate, and $\epsilon$ is the clipping threshold for stable optimization.

For reward computation, we extract the final answer enclosed by the \texttt{<answer>...</answer>} tags from each rollout and use \textbf{Qwen3.5-27B} as an LLM judge to evaluate its correctness against the normalized reference answer. The resulting judgment is used as the reward signal for policy optimization.

\textbf{Multi-Teacher On-Policy Distillation.}
After obtaining the complementary RL experts, we integrate their specialized capabilities back into the unified SFT model through \textbf{multi-teacher on-policy distillation (MOPD)}. The student is initialized from the SFT generalist and generates responses using its current policy. For expert routing, each training sample is assigned to the corresponding Real-World or Analytical expert according to its reasoning group. Rather than generating independent teacher trajectories, the selected expert evaluates the prefixes actually visited by the student and provides token-level distributional supervision on these on-policy states.

Specifically, for each student prefix $s_t=(x,y_{<t})$, the routed expert $\pi_{\phi_g}$ provides its top-$K$ token probabilities. The student probabilities are evaluated on the same teacher-selected token support, allowing the student to absorb the distributional preferences of the expert. We consider both forward- and reverse-KL objectives on the teacher-selected top-$K$ token support, with the core forms:
\[
\mathcal{L}_{\rm FKL}
=
\frac{1}{|y|}
\sum_{t=1}^{|y|}
\sum_{v\in\mathcal{T}_t^g}
\pi_{\phi_g}(v\mid s_t)
\log
\frac{\pi_{\phi_g}(v\mid s_t)}{\pi_{\theta}(v\mid s_t)}
\;,\;
\mathcal{L}_{\rm RKL}
=
\frac{1}{|y|}
\sum_{t=1}^{|y|}
\sum_{v\in\mathcal{T}_t^g}
\pi_{\theta}(v\mid s_t)
\log
\frac{\pi_{\theta}(v\mid s_t)}{\pi_{\phi_g}(v\mid s_t)}.
\]
Here, $g\in\{\text{real-world},\mathrm{analytical}\}$ denotes the reasoning group used for expert routing, and $\mathcal{T}_t^g$ denotes the top-$K$ token support selected by the routed expert $\pi_{\phi_g}$ at prefix $s_t$. Following MOPD~\cite{ma2026mopd}, Reverse-KL further applies a correction for top-$K$ truncation bias. We investigate both forward- and reverse-KL objectives to study how different distribution-matching directions affect the integration of complementary expert capabilities into the final MMVistaReason models.

\section{Experiments}

\subsection{Experimental Setup}

\textbf{Supervised Fine-Tuning.}
We initialize the model from Qwen3.5-Base and perform supervised fine-tuning on \textbf{MMVistaReason-SFT-528K} using LLaMA-Factory \cite{llamafactory}. All training responses follow the \texttt{<think>...</think>} and \texttt{<answer>...</answer>} format to encourage structured reasoning.

\textbf{Reinforcement Learning.}
We train complementary RL experts with VeRL \cite{verl} on scale-specific subsets of \textbf{MMVistaReason-RL-63K}, corresponding to Real-World and Analytical Reasoning. We optimize both experts with GSPO using a global batch size of 256 and sample 8 rollouts per training example.

\textbf{On-Policy Distillation.}
We perform MOPD with VeRL \cite{verl} to integrate the Real-World and Analytical experts into a unified model. The student is initialized from the SFT model, and each sample is routed to its corresponding expert for teacher-top-$K$ supervision on student trajectories, with $K=32$.

\textbf{Evaluation and Benchmarks.}
We evaluate all models with VLMEvalKit \cite{duan2024vlmevalkit} on diverse multimodal benchmarks. We use greedy decoding with temperature set to 0 and set the maximum generation length to 16,384 tokens. We report both domain-level scores and overall average performance.

To ensure a comprehensive assessment, we evaluate our model across 15 multimodal benchmarks spanning two key reasoning categories:
\begin{itemize}
    \item \textbf{Real-World Visual Reasoning:} covers diverse real-world visual understanding capabilities, including GUI interaction, general VQA, spatial reasoning, visual perception, and counting. We evaluate on OSWorld-G \cite{xie2026scaling}, CV-Bench-2D \cite{tong2024cambrian}, CV-Bench-3D \cite{tong2024cambrian}, MMBench-EN \cite{liu2024mmbench}, RealWorldQA \cite{xai2024grok15v} and CountBenchQA \cite{paiss2023teaching}.

    \item \textbf{Analytical Visual Reasoning:} focuses on structured and knowledge-intensive reasoning, including science, mathematics, logic/puzzle, and chart/document reasoning. We evaluate on Scientists' First Exam (SFE) \cite{zhou2026scientists}, MMMU$_{\mathrm{val}}$ \cite{yue2024mmmu}, ScienceQA \cite{lu2022learn}, MathVista$_{\mathrm{mini}}$ \cite{lu2024mathvista}, MathVerse$_{\mathrm{mini}}$ \cite{zhang2024mathverse}, LogicVista$_{\mathrm{test}}$ \cite{xiao2024logicvista}, VisuLogic$_{\mathrm{test}}$ \cite{xu2026visulogic}, ChartQA \cite{masry2022chartqa}, and CharXiv$_{\mathrm{reas.}}$ \cite{wang2024charxiv}.
\end{itemize}

\textbf{Baselines.} Our baselines fall into three main categories for comprehensive comparison: 
(1) \textbf{Closed-source VLMs}, including Gemini-3-Flash \cite{google2025gemini3flash} and GPT-5.1 \cite{openai2025gpt51}; 
(2) \textbf{Open-weight VLMs} (Instruct mode), including Qwen3.5-9B \cite{qwen2026qwen35_9b}, Qwen3.5-35B-A3B \cite{qwen2026qwen35_35b_a3b}, InternVL3.5-30B-A3B \cite{wang2025internvl3}, and InternVL3.5-241B-A28B \cite{wang2025internvl3}; (3) \textbf{Open-source Reasoning VLMs}, including OpenMMReasoner (OMR)-7B \cite{zhang2026openmmreasoner}, MMFineReason (MFR)-4B \cite{lin2026mmfinereason}, and MMFineReason (MFR)-8B \cite{lin2026mmfinereason}.

\subsection{Main Results}

Table~\ref{tab:main_results} compares \textbf{MMVistaReason} models with representative closed-source, open-weight, and open-source VLMs. MMVistaReason achieves strong performance at both scales, with overall averages of \textbf{72.8} and \textbf{74.4} for the 4B and 9B models, respectively. Despite its compact scale, MMVistaReason-4B outperforms MFR-8B by 1.9 points and Qwen3.5-9B (Instruct) by 0.7 points. Scaling to 9B further improves performance, surpassing the much larger Qwen3.5-35B-A3B (Instruct) by 0.3 points and approaching GPT-5.1, demonstrating strong parameter efficiency and scalability.

\begin{table*}[t]
\centering
\caption{Comparison of MMVistaReason (MVR) models with representative closed-source, open-weight, and open-source VLMs across diverse multimodal benchmarks.}
\label{tab:main_results}
\setlength{\tabcolsep}{3.2pt}
\renewcommand{\arraystretch}{1.08}
\fontsize{7.5pt}{8.5pt}\selectfont

\resizebox{\textwidth}{!}{
\begin{tabular}{lccccccccccc}
\toprule
\multirow{2}{*}{\textbf{Benchmarks}}
& \multicolumn{2}{c}{\textbf{Closed-source VLMs}}
& \multicolumn{4}{c}{\textbf{Open-weight VLMs}}
& \multicolumn{3}{c}{\textbf{Open-source VLMs}}
& \multicolumn{2}{c}{\textbf{Ours}} \\
\cmidrule(lr){2-3}
\cmidrule(lr){4-7}
\cmidrule(lr){8-10}
\cmidrule(lr){11-12}

& \makecell{\textbf{Gemini-3}\\\textbf{Flash}}
& \makecell{\textbf{GPT-5.1}}
& \makecell{\textbf{Qwen3.5-9B}\\\textbf{(Instruct)}}
& \makecell{\textbf{Qwen3.5-35B}\\\textbf{A3B (Instruct)}}
& \makecell{\textbf{InternVL3.5}\\\textbf{30B-A3B}}
& \makecell{\textbf{InternVL3.5}\\\textbf{241B-A28B}}
& \makecell{\textbf{OMR}\\\textbf{7B}}
& \makecell{\textbf{MFR}\\\textbf{4B}}
& \makecell{\textbf{MFR}\\\textbf{8B}}
& \cellcolor{blue!8}\makecell{\textbf{MVR}\\\textbf{4B}}
& \cellcolor{blue!8}\makecell{\textbf{MVR}\\\textbf{9B}} \\

\midrule

OSWorld-G
& 65.4 & 62.6 & 60.4 & 62.8 & 45.7 & 54.9 & 32.8 & 41.4 & 48.3
& \cellcolor{blue!8} 58.5
& \cellcolor{blue!8} 64.8 \\

CV-Bench-2D
& 84.6 & 83.0 & 82.9 & 79.4 & 80.3 & 81.2 & 76.9 & 79.2 & 80.2
& \cellcolor{blue!8} 82.4
& \cellcolor{blue!8} 83.2 \\

CV-Bench-3D
& 92.8 & 91.6 & 91.3 & 93.1 & 87.8 & 89.3 & 84.0 & 89.1 & 90.8
& \cellcolor{blue!8} 91.8
& \cellcolor{blue!8} 92.4 \\

MMBench-EN
& 91.5 & 89.6 & 90.5 & 92.1 & 85.2 & 87.8 & 88.3 & 88.8 & 89.5
& \cellcolor{blue!8} 90.6
& \cellcolor{blue!8} 92.2 \\

RealWorldQA
& 80.9 & 79.1 & 76.8 & 76.7 & 72.3 & 75.2 & 68.8 & 74.6 & 75.2
& \cellcolor{blue!8} 77.4
& \cellcolor{blue!8} 77.4 \\

CountBenchQA
& 95.3 & 91.8 & 89.5 & 92.1 & 84.7 & 90.4 & 81.9 & 87.1 & 88.1
& \cellcolor{blue!8} 91.5
& \cellcolor{blue!8} 92.8 \\

\rowcolor{blue!8}
\textbf{Real-World Avg.}
& 85.1 & 83.0 & 81.9 & 82.7 & 76.0 & 79.8 & 72.1 & 76.7 & 78.7
& 82.0
& 83.8 \\

\midrule

SFE
& 30.5 & 27.1 & 18.8 & 22.3 & 20.7 & 19.6 & 11.2 & 15.7 & 17.0
& \cellcolor{blue!8} 19.8
& \cellcolor{blue!8} 20.7 \\

MMMU$_{\mathrm{val}}$
& 77.2 & 77.6 & 69.2 & 76.1 & 69.4 & 73.3 & 57.6 & 69.6 & 71.2
& \cellcolor{blue!8} 71.8
& \cellcolor{blue!8} 73.6 \\

ScienceQA
& 98.7 & 96.9 & 97.1 & 98.2 & 97.7 & 98.2 & 96.9 & 96.3 & 97.2
& \cellcolor{blue!8} 96.8
& \cellcolor{blue!8} 97.4 \\

MathVista$_{\mathrm{mini}}$
& 85.4 & 83.9 & 82.1 & 84.2 & 75.6 & 82.7 & 79.3 & 80.5 & 82.1
& \cellcolor{blue!8} 82.4
& \cellcolor{blue!8} 83.6 \\

MathVerse$_{\mathrm{mini}}$
& 83.3 & 80.7 & 76.9 & 80.6 & 61.8 & 68.5 & 63.7 & 78.7 & 80.4
& \cellcolor{blue!8} 79.7
& \cellcolor{blue!8} 82.4 \\

LogicVista$_{\mathrm{test}}$
& 70.2 & 68.2 & 65.1 & 68.5 & 53.9 & 66.7 & 51.9 & 66.9 & 68.5
& \cellcolor{blue!8} 66.4
& \cellcolor{blue!8} 68.5 \\

VisuLogic$_{\mathrm{test}}$
& 34.7 & 31.9 & 28.9 & 30.1 & 28.8 & 29.1 & 26.0 & 29.4 & 29.8
& \cellcolor{blue!8} 29.0
& \cellcolor{blue!8} 30.2 \\

ChartQA
& 88.0 & 85.7 & 85.1 & 86.3 & 82.1 & 83.4 & 84.1 & 84.2 & 85.1
& \cellcolor{blue!8} 86.2
& \cellcolor{blue!8} 87.6 \\

CharXiv$_{\mathrm{reas.}}$
& 75.6 & 69.5 & 67.2 & 69.6 & 46.7 & 47.6 & 49.1 & 58.2 & 60.8
& \cellcolor{blue!8} 66.9
& \cellcolor{blue!8} 68.8 \\

\rowcolor{blue!8}
\textbf{Analytical Avg.}
& 71.5 & 69.1 & 65.6 & 68.4 & 59.6 & 63.2 & 57.8 & 64.4 & 65.8
& 66.6
& 68.1 \\

\midrule

\textbf{Overall Avg.}
& 76.9 & 74.6 & 72.1 & 74.1 & 66.2 & 69.9 & 63.5 & 69.3 & 70.9
& \cellcolor{blue!8} 72.8
& \cellcolor{blue!8} 74.4 \\

\bottomrule
\end{tabular}
}
\end{table*}

\textbf{Strong Real-World Visual Reasoning.}
MVR-4B achieves a Real-World average of \textbf{82.0}, outperforming MFR-8B by 3.3 points, with notable gains on OSWorld-G (\textbf{58.5}, +10.2) and CountBenchQA (\textbf{91.5}, +3.4). Scaling to 9B further raises the average to \textbf{83.8}, surpassing Qwen3.5-35B-A3B (Instruct) by 1.1 points, with particularly strong results on OSWorld-G (\textbf{64.8}) and CountBenchQA (\textbf{92.8}). These results demonstrate strong and scalable capabilities across diverse real-world visual reasoning tasks.

\textbf{Competitive Analytical Visual Reasoning.}
MVR-4B achieves an Analytical average of \textbf{66.6}, outperforming Qwen3.5-9B (Instruct) and MFR-8B by 1.0 and 0.8 points, respectively. Scaling to 9B further improves the average to \textbf{68.1}, approaching the much larger Qwen3.5-35B-A3B (Instruct). Notably, MVR-9B achieves \textbf{82.4} on MathVerse$_{\mathrm{mini}}$, surpassing the 35B model and GPT-5.1, while reaching \textbf{87.6} on ChartQA. These results demonstrate strong analytical reasoning with high parameter efficiency.

\textbf{Broad Generalization across Diverse Multimodal Scenarios.}
Across 15 diverse multimodal benchmarks, MMVistaReason demonstrates broad coverage across both Real-World and Analytical scenarios. Its \textbf{consistent and balanced} performance shows that the gains are not confined to a single domain, but extend across perception, spatial understanding, science, mathematics, logic, and chart reasoning. Notably, both 4B and 9B models maintain strong performance across heterogeneous tasks, highlighting the effectiveness and scalability of our unified post-training recipe.

\subsection{Performance across Training Stages}

Table~\ref{tab:training_stages} compares different training stages across 4B and 9B model scales. Overall, our training recipe consistently improves performance across diverse multimodal benchmarks. At both scales, the SFT models surpass their corresponding Qwen3.5 (Instruct) counterparts. The final MMVistaReason-4B achieves an overall average of \textbf{72.8}, outperforming Qwen3.5-4B (Thinking), while MMVistaReason-9B reaches \textbf{74.4}, performing on par with and slightly surpassing Qwen3.5-9B (Thinking).

\begin{table*}[t]
\centering
\caption{Performance comparison across different model scales and training stages on multimodal benchmarks. The best result within each model scale is highlighted in bold. Inst. and Think. denote the instruct and thinking modes of the Qwen3.5 models, respectively.}
\label{tab:training_stages}

\setlength{\tabcolsep}{3.4pt}
\renewcommand{\arraystretch}{1.03}
\fontsize{3.6pt}{4.2pt}\selectfont

\setlength{\heavyrulewidth}{0.38pt}
\setlength{\lightrulewidth}{0.22pt}
\setlength{\cmidrulewidth}{0.20pt}
\setlength{\aboverulesep}{0.30ex}
\setlength{\belowrulesep}{0.30ex}
\setlength{\abovetopsep}{0pt}
\setlength{\belowbottomsep}{0pt}

\resizebox{\textwidth}{!}{
\begin{tabular}{
l
ccc>{\columncolor{blue!8}}c>{\columncolor{blue!8}}c
ccc>{\columncolor{blue!8}}c>{\columncolor{blue!8}}c
}
\toprule

\multirow{2}{*}{\textbf{Benchmark}}
& \multicolumn{5}{c}{\textbf{4B Models}}
& \multicolumn{5}{c}{\textbf{9B Models}} \\

\cmidrule(l{3pt}r{3pt}){2-6}
\cmidrule(l{3pt}r{3pt}){7-11}

& \textbf{Base}
& \textbf{Inst.}
& \textbf{Think.}
& \textbf{SFT}
& \textbf{MOPD}
& \textbf{Base}
& \textbf{Inst.}
& \textbf{Think.}
& \textbf{SFT}
& \textbf{MOPD} \\

\midrule

OSWorld-G
& 41.5
& 52.0
& 54.2
& 54.0
& \textbf{58.5}
& 43.2
& 60.4
& 62.6
& 59.9
& \textbf{64.8} \\

CV-Bench-2D
& 80.9
& \textbf{82.7}
& 82.1
& 80.8
& 82.4
& 81.8
& 82.9
& 83.1
& 80.3
& \textbf{83.2} \\

CV-Bench-3D
& 91.2
& 90.5
& \textbf{91.9}
& 90.6
& 91.8
& 91.6
& 91.3
& \textbf{92.6}
& 91.8
& 92.4 \\

MMBench-EN
& 89.8
& 90.2
& 90.0
& 90.1
& \textbf{90.6}
& 91.0
& 90.5
& 90.8
& 91.7
& \textbf{92.2} \\

RealWorldQA
& 75.2
& 75.0
& \textbf{77.9}
& 73.9
& 77.4
& 75.3
& 76.8
& \textbf{79.0}
& 75.2
& 77.4 \\

CountBenchQA
& 84.7
& 87.3
& \textbf{92.6}
& 88.7
& 91.5
& 87.3
& 89.5
& \textbf{94.3}
& 90.7
& 92.8 \\

\rowcolor{blue!8}
\textbf{Real-World Avg.}
& 77.2
& 79.6
& 81.4
& 79.7
& \textbf{82.0}
& 78.4
& 81.9
& 83.7
& 81.6
& \textbf{83.8} \\

\midrule

SFE
& 16.6
& \textbf{20.1}
& 19.6
& 17.8
& 19.8
& 16.9
& 18.8
& 20.1
& 18.2
& \textbf{20.7} \\

MMMU$_{\mathrm{val}}$
& 63.9
& 67.1
& 70.0
& 70.7
& \textbf{71.8}
& 67.3
& 69.2
& 72.8
& 71.4
& \textbf{73.6} \\

ScienceQA
& 94.2
& 95.8
& \textbf{97.1}
& 96.2
& 96.8
& 95.3
& 97.1
& 97.2
& 96.9
& \textbf{97.4} \\

MathVista$_{\mathrm{mini}}$
& 75.4
& 81.5
& \textbf{83.5}
& 80.4
& 82.4
& 79.2
& 82.1
& \textbf{83.8}
& 82.1
& 83.6 \\

MathVerse$_{\mathrm{mini}}$
& 66.3
& 73.3
& 76.4
& 78.3
& \textbf{79.7}
& 72.0
& 76.9
& 81.9
& 80.9
& \textbf{82.4} \\

LogicVista$_{\mathrm{test}}$
& 58.1
& 62.9
& \textbf{68.5}
& 65.1
& 66.4
& 61.7
& 65.1
& \textbf{69.4}
& 66.0
& 68.5 \\

VisuLogic$_{\mathrm{test}}$
& 21.3
& 27.8
& \textbf{29.2}
& 28.1
& 29.0
& 22.5
& 28.9
& \textbf{31.2}
& 28.8
& 30.2 \\

ChartQA
& 82.6
& 85.2
& 85.1
& 83.9
& \textbf{86.2}
& 85.6
& 85.1
& 87.2
& 85.1
& \textbf{87.6} \\

CharXiv$_{\mathrm{reas.}}$
& 57.9
& 65.1
& 66.3
& 63.6
& \textbf{66.9}
& 62.8
& 67.2
& \textbf{69.0}
& 65.6
& 68.8 \\

\rowcolor{blue!8}
\textbf{Analytical Avg.}
& 59.6
& 64.3
& 66.2
& 64.9
& \textbf{66.6}
& 62.6
& 65.6
& \textbf{68.1}
& 66.1
& \textbf{68.1} \\

\midrule

\textbf{Overall Avg.}
& 66.6
& 70.4
& 72.3
& 70.8
& \textbf{72.8}
& 68.9
& 72.1
& 74.3
& 72.3
& \textbf{74.4} \\

\bottomrule
\end{tabular}
}

\end{table*}

\textbf{SFT Establishes a Strong Reasoning Foundation.}
Starting from Qwen3.5-Base models, our SFT models improve the overall average from 66.6 to 70.8 at 4B and from 68.9 to 72.3 at 9B, both surpassing their corresponding Qwen3.5-Instruct models. At 4B, the gains over the Instruct model are pronounced on reasoning-intensive benchmarks, including MMMU$_{\mathrm{val}}$ (67.1$\rightarrow$70.7), MathVerse$_{\mathrm{mini}}$ (73.3$\rightarrow$78.3), and LogicVista$_{\mathrm{test}}$ (62.9$\rightarrow$65.1), with similar trends observed at 9B. These results suggest that our curated SFT trajectories consistently strengthen analytical reasoning and generalize well across model scales, providing a strong foundation for subsequent reinforcement learning.

\textbf{MOPD Further Integrates Complementary Capabilities.}
After RL expert specialization and MOPD integration, MMVistaReason improves the overall average from 70.8 to \textbf{72.8} at 4B and from 72.3 to \textbf{74.4} at 9B. The gains are consistent across both reasoning groups, with Real-World averages improving from 79.7 to 82.0 and 81.6 to 83.8, and Analytical averages from 64.9 to 66.6 and 66.1 to 68.1, respectively. Notable gains over the SFT models appear on OSWorld-G (+4.5/+4.9), RealWorldQA (+3.5/+2.2), and CharXiv$_{\mathrm{reas.}}$ (+3.3/+3.2) at 4B/9B. These results indicate that expert specialization followed by MOPD effectively consolidates complementary capabilities into stronger unified models across scales.

\subsection{Further Analysis}
To further examine the effectiveness and robustness of our proposed framework, we conduct a series of in-depth analyses from multiple perspectives. These studies provide additional insights into the key design choices and training behaviors of MMVistaReason.

\begin{figure*}[t]
    \centering
    \includegraphics[width=0.95\textwidth]{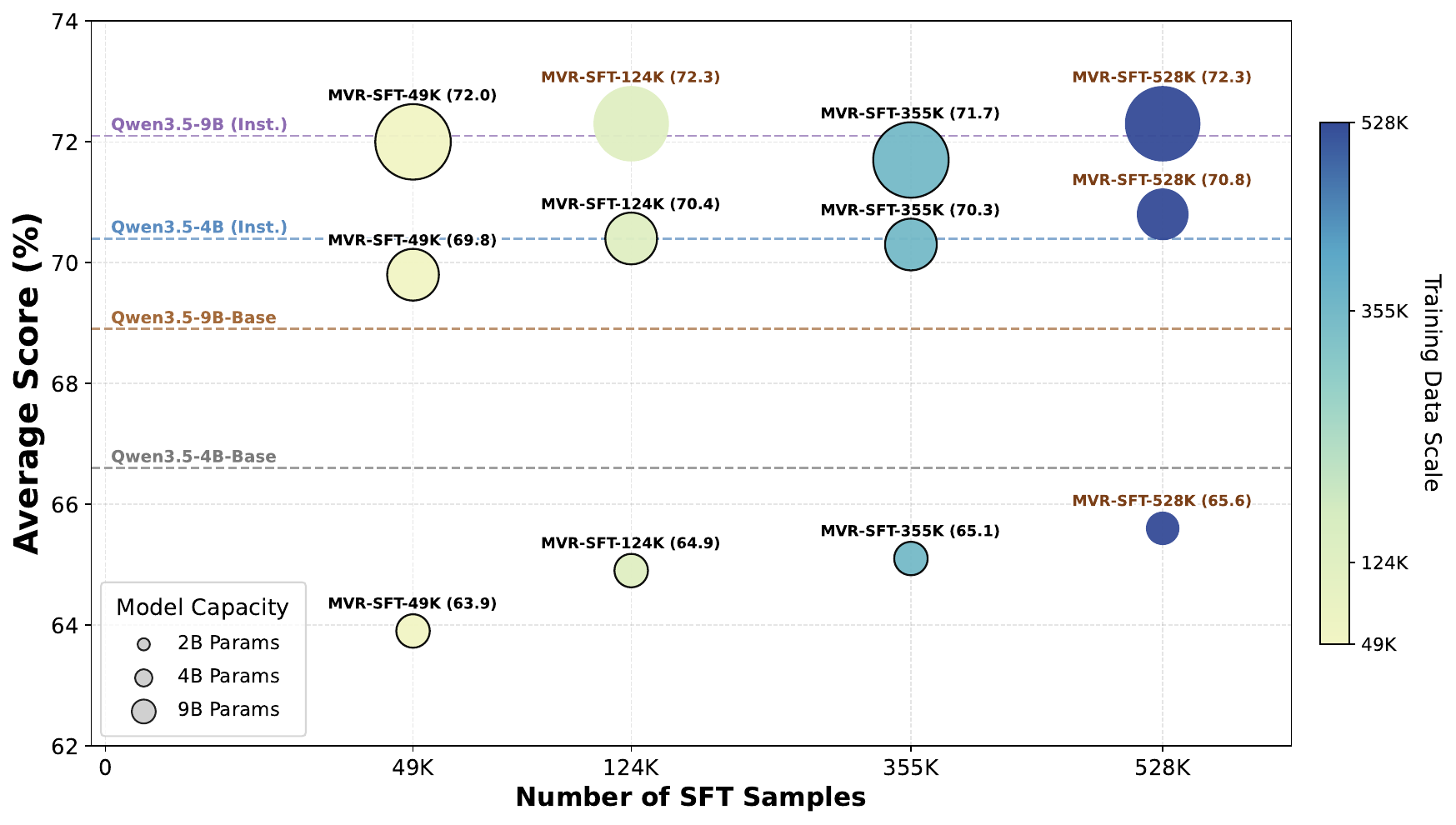}
    \caption{Data scale and performance trade-off across different student model capacities.}
    \label{fig:data_scale}
\end{figure*}

\textbf{Data Scale and Performance Trade-off.} Figure~\ref{fig:data_scale} investigates the relationship among SFT data scale, supervision difficulty, and student model capacity. To analyze the effect of model capacity, we fine-tune Qwen3.5-2B-Base, Qwen3.5-4B-Base, and Qwen3.5-9B-Base on different SFT subsets under identical training settings. We compare MVR-SFT-49K, MVR-SFT-124K, and MVR-SFT-355K, which are constructed from different teachers in the cascade, together with the full mixed dataset MVR-SFT-528K. These subsets differ not only in scale, but also in teacher source and sample difficulty, allowing us to further examine how data scale and supervision difficulty interact with student model capacity.

For smaller models, broader supervision spanning multiple difficulty levels brings consistent gains. The 2B model improves from 63.9 on the 49K subset to 65.6 on the full 528K mixture, while the 4B model improves from 69.8 to 70.8, surpassing the Qwen3.5-4B-Instruct baseline of 70.4. Notably, the 49K subset is generated by the strongest 122B teacher and consists of more difficult samples that cannot be solved by the preceding teachers. However, using this subset alone yields relatively weak performance for both the 2B and 4B students. This suggests that high-quality yet overly difficult supervision may not be fully absorbed by smaller students; compared with training only on the hardest reasoning trajectories, the full mixture with multiple difficulty levels is better suited to compact models.

In contrast, the 9B model exhibits a non-monotonic trend. The 124K subset generated by the 27B teacher achieves an overall score of \textbf{72.3}, matching the full 528K mixture while outperforming the 355K subset at 71.7 and the Qwen3.5-9B (Instruct) baseline at 72.1. In particular, the 355K subset is generated by a teacher of the same 9B scale. Despite being the largest individual subset, it provides only limited gains for the 9B student, suggesting that samples already solvable by a same-capacity teacher may offer insufficient additional supervision. By contrast, moderately harder supervision from the stronger 27B teacher enables the 9B student to match full-data performance with only 124K samples, highlighting the importance of supervision difficulty over data scale for larger students.

Overall, these results show that more SFT data is not always better; its effectiveness depends critically on the match between supervision difficulty and student model capacity, rather than on scaling data volume alone. This further highlights the value of cascaded data construction in providing difficulty-stratified supervision that better matches students of different capacities.

\begin{figure*}[t]
    \centering
    \includegraphics[width=1\textwidth]{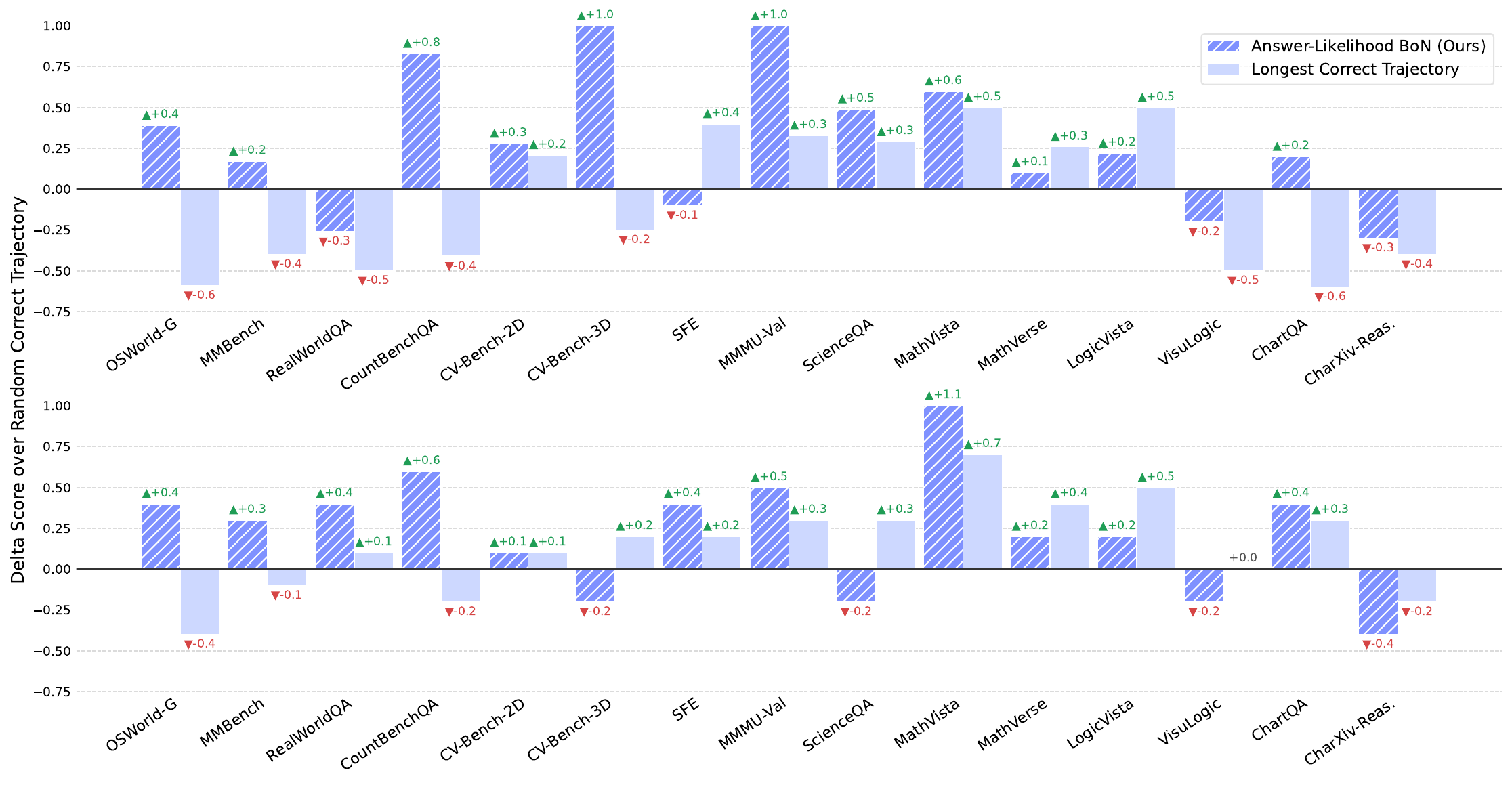}
    \caption{Benchmark-wise score changes of different reasoning trajectory selection strategies over random correct trajectory selection for 4B (top) and 9B (bottom) student models.}
    \label{fig:trajectory_selection}
\end{figure*}

\textbf{Effect of Reasoning Trajectory Selection.}
We further investigate the impact of reasoning trajectory selection across 4B and 9B student models. For each training sample, multiple teacher-generated trajectories may reach the same correct final answer, while their supervision quality can vary substantially. To isolate the effect of trajectory selection, we fix both the training samples and candidate trajectory pool, and compare three strategies: random correct trajectory selection, longest correct trajectory selection, and our answer-likelihood Best-of-N selection.

Figure~\ref{fig:trajectory_selection} reports the benchmark-wise score changes over random selection. Answer-likelihood Best-of-N yields positive average gains at both model scales. At 4B, notable improvements are observed on CV-Bench-3D (+1.0), MMMU$_{\mathrm{val}}$ (+1.0), and CountBenchQA (+0.8), while at 9B, the largest gains appear on MathVista$_{\mathrm{mini}}$ (+1.1), CountBenchQA (+0.6), and MMMU$_{\mathrm{val}}$ (+0.5). Although slight degradation occurs on a few benchmarks, the overall trend indicates that answer likelihood serves as an effective proxy for identifying higher-quality reasoning trajectories. In contrast, selecting the longest correct trajectory is less stable, particularly at 4B. Longer trajectories can introduce redundancy or noise and therefore do not consistently provide better supervision. Nevertheless, the longest-trajectory strategy outperforms our selection on several benchmarks, especially at 9B, suggesting that tasks requiring more extensive multi-step reasoning may still benefit from longer reasoning traces. Thus, trajectory length can be useful for particular tasks, but is less reliable as a general selection criterion.

We observe that trajectory selection has a smaller impact on the 9B student. The average gain of answer-likelihood Best-of-N over random selection decreases from +0.29 at 4B to +0.24 at 9B, while the longest-trajectory strategy becomes more stable. This suggests that larger students are more robust to variations in trajectory quality, whereas smaller students benefit more from careful trajectory selection. Overall, these results validate answer-likelihood Best-of-N as an effective trajectory selection strategy and highlight the interaction between supervision quality and student model capacity.

\textbf{Performance of Complementary RL Experts.}
Table~\ref{tab:rl_experts} examines whether the two RL experts develop complementary capabilities across model scales. Real-world benchmarks emphasize perception, spatial understanding, counting, and visual grounding, whereas analytical benchmarks focus on structured, knowledge-intensive, and symbolic reasoning. We therefore train the Real-World and Analytical RL Experts from corresponding MVR-SFT models and evaluate their specialization patterns.

The Real-World Expert strengthens real-world visual reasoning across model scales. Its Real-World Avg. improves from 79.7 to 82.7 at 4B and from 81.6 to 84.5 at 9B, with gains across all six real-world benchmarks. Representative improvements include OSWorld-G (54.0$\rightarrow$59.5 / 59.9$\rightarrow$66.3) and RealWorldQA (73.9$\rightarrow$78.6 / 75.2$\rightarrow$77.9). Meanwhile, its Analytical Avg. decreases slightly, from 64.9 to 64.3 at 4B and from 66.1 to 66.0 at 9B, suggesting weaker cross-domain interference at the larger scale.

The Analytical Expert also exhibits clear specialization across model scales. Its Analytical Avg. increases from 64.9 to 66.6 at 4B and from 66.1 to 68.2 at 9B, with gains across all nine analytical benchmarks. Representative improvements include SFE (17.8$\rightarrow$19.6 / 18.2$\rightarrow$20.6) and CharXiv$_{\mathrm{reas.}}$ (63.6$\rightarrow$66.7 / 65.6$\rightarrow$69.3) at 4B/9B. In contrast, its Real-World Avg. decreases slightly from 79.7 to 79.2 at 4B but remains unchanged at 81.6 for 9B, showing that stronger analytical specialization can be achieved without noticeably sacrificing overall real-world performance at the larger scale.

Overall, the two experts exhibit clear complementary specialization across model scales: the Real-World Expert strengthens perceptual and real-world capabilities, whereas the Analytical Expert improves structured analytical reasoning. Their domain-specific gains and reduced cross-group interference at larger scale validate our domain-grouped RL design and motivate multi-teacher integration.

\begin{table*}[t]
\centering
\caption{
Benchmark-wise and domain-level performance of complementary RL experts across different model scales.
RW and Ana. denote Real-World and Analytical experts, respectively.
Benchmark names are abbreviated due to space constraints.
Green and red backgrounds indicate gains and drops over the corresponding SFT model.
The $\Delta$ rows report the improvement of MOPD over SFT.
}
\label{tab:rl_experts}

\setlength{\tabcolsep}{2.6pt}
\renewcommand{\arraystretch}{1.03}
\fontsize{6.6pt}{7.4pt}\selectfont

\resizebox{\textwidth}{!}{
\begin{tabular}{>{\centering\arraybackslash}p{0.4cm} c *{17}{c}}
\toprule

\multirow{2}{*}{\textbf{Scale}}
& \multirow{2}{*}{\textbf{Stage}}
& \multicolumn{7}{c}{\textbf{Real-World Visual Reasoning}}
& \multicolumn{10}{c}{\textbf{Analytical Visual Reasoning}} \\

\cmidrule(lr){3-9}
\cmidrule(lr){10-19}

&
& OSW & CV2D & CV3D & MMB & RWA & Count & \textbf{Avg.}
& SFE & MMMU & SQA & MVis & MVer & LVis & VLog & CQA & CharXiv & \textbf{Avg.} \\

\midrule


\multirow{5}{*}{\textbf{4B}}
& SFT
& 54.0
& 80.8
& 90.6
& 90.1
& 73.9
& 88.7
& 79.7
& 17.8
& 70.7
& 96.2
& 80.4
& 78.3
& 65.1
& 28.1
& 83.9
& 63.6
& 64.9 \\

& RW Expert
& \poscell{35}{\textbf{59.5}}
& \poscell{25}{\textbf{82.9}}
& \poscell{22}{\textbf{92.3}}
& \poscell{18}{\textbf{91.2}}
& \poscell{32}{\textbf{78.6}}
& \poscell{28}{91.4}
& \poscell{30}{\textbf{82.7}}
& \negcell{12}{17.0}
& \negcell{30}{67.6}
& \negcell{12}{95.9}
& \poscell{6}{80.5}
& \negcell{18}{77.1}
& \negcell{10}{64.9}
& \negcell{12}{27.8}
& \negcell{15}{83.1}
& \poscell{18}{64.7}
& \negcell{12}{64.3} \\

& Ana. Expert
& \negcell{10}{53.7}
& \negcell{24}{78.9}
& 90.6
& 90.1
& \poscell{14}{74.3}
& \negcell{15}{87.7}
& \negcell{12}{79.2}
& \poscell{28}{19.6}
& \poscell{16}{71.4}
& \poscell{20}{\textbf{97.2}}
& \poscell{26}{\textbf{82.5}}
& \poscell{20}{\textbf{79.8}}
& \poscell{22}{\textbf{66.7}}
& \poscell{18}{29.1}
& \poscell{30}{\textbf{86.7}}
& \poscell{34}{66.7}
& \poscell{28}{\textbf{66.6}} \\

& \cellcolor{blue!8}\textbf{MOPD}
& \cellcolor{blue!8} 58.5
& \cellcolor{blue!8} 82.4
& \cellcolor{blue!8} 91.8
& \cellcolor{blue!8} 90.6
& \cellcolor{blue!8} 77.4
& \cellcolor{blue!8} \textbf{91.5}
& \cellcolor{blue!8} 82.0
& \cellcolor{blue!8} \textbf{19.8}
& \cellcolor{blue!8} \textbf{71.8}
& \cellcolor{blue!8} 96.8
& \cellcolor{blue!8} 82.4
& \cellcolor{blue!8} 79.7
& \cellcolor{blue!8} 66.4
& \cellcolor{blue!8} \textbf{29.0}
& \cellcolor{blue!8} 86.2
& \cellcolor{blue!8} \textbf{66.9}
& \cellcolor{blue!8} \textbf{66.6} \\

& \cellcolor{blue!8}\scriptsize$\Delta$ vs. SFT
& \cellcolor{blue!8}\textcolor{green!50!black}{\scriptsize$\uparrow$4.5}
& \cellcolor{blue!8}\textcolor{green!50!black}{\scriptsize$\uparrow$1.6}
& \cellcolor{blue!8}\textcolor{green!50!black}{\scriptsize$\uparrow$1.2}
& \cellcolor{blue!8}\textcolor{green!50!black}{\scriptsize$\uparrow$0.5}
& \cellcolor{blue!8}\textcolor{green!50!black}{\scriptsize$\uparrow$3.5}
& \cellcolor{blue!8}\textcolor{green!50!black}{\scriptsize$\uparrow$2.8}
& \cellcolor{blue!8}\textcolor{green!50!black}{\scriptsize$\uparrow$\textbf{2.3}}
& \cellcolor{blue!8}\textcolor{green!50!black}{\scriptsize$\uparrow$2.0}
& \cellcolor{blue!8}\textcolor{green!50!black}{\scriptsize$\uparrow$1.1}
& \cellcolor{blue!8}\textcolor{green!50!black}{\scriptsize$\uparrow$0.6}
& \cellcolor{blue!8}\textcolor{green!50!black}{\scriptsize$\uparrow$2.0}
& \cellcolor{blue!8}\textcolor{green!50!black}{\scriptsize$\uparrow$1.4}
& \cellcolor{blue!8}\textcolor{green!50!black}{\scriptsize$\uparrow$1.3}
& \cellcolor{blue!8}\textcolor{green!50!black}{\scriptsize$\uparrow$0.9}
& \cellcolor{blue!8}\textcolor{green!50!black}{\scriptsize$\uparrow$2.3}
& \cellcolor{blue!8}\textcolor{green!50!black}{\scriptsize$\uparrow$3.3}
& \cellcolor{blue!8}\textcolor{green!50!black}{\scriptsize$\uparrow$\textbf{1.7}} \\

\midrule


\multirow{5}{*}{\textbf{9B}}
& SFT
& 59.9
& 80.3
& 91.8
& 91.7
& 75.2
& 90.7
& 81.6
& 18.2
& 71.4
& 96.9
& 82.1
& 80.9
& 66.0
& 28.8
& 85.1
& 65.6
& 66.1 \\

& RW Expert
& \poscell{35}{\textbf{66.3}}
& \poscell{32}{\textbf{84.4}}
& \poscell{18}{\textbf{92.8}}
& \poscell{18}{\textbf{92.7}}
& \poscell{28}{\textbf{77.9}}
& \poscell{24}{92.6}
& \poscell{30}{\textbf{84.5}}
& \negcell{10}{17.8}
& \negcell{16}{70.6}
& \poscell{6}{97.0}
& \poscell{8}{82.2}
& \negcell{18}{80.0}
& \poscell{20}{66.8}
& \negcell{14}{28.0}
& \negcell{8}{84.7}
& \poscell{24}{67.4}
& \negcell{5}{66.0} \\

& Ana. Expert
& \poscell{18}{61.1}
& \poscell{12}{81.1}
& \negcell{6}{91.7}
& \negcell{5}{91.5}
& \negcell{10}{74.6}
& \negcell{15}{89.7}
& 81.6
& \poscell{30}{20.6}
& \poscell{24}{73.3}
& \poscell{12}{\textbf{97.5}}
& \poscell{30}{\textbf{84.5}}
& \poscell{24}{\textbf{82.7}}
& \poscell{26}{68.0}
& \poscell{20}{30.1}
& \poscell{28}{87.4}
& \poscell{35}{\textbf{69.3}}
& \poscell{30}{\textbf{68.2}} \\

& \cellcolor{blue!8}\textbf{MOPD}
& \cellcolor{blue!8} 64.8
& \cellcolor{blue!8} 83.2
& \cellcolor{blue!8} 92.4
& \cellcolor{blue!8} 92.2
& \cellcolor{blue!8} 77.4
& \cellcolor{blue!8} \textbf{92.8}
& \cellcolor{blue!8} 83.8
& \cellcolor{blue!8} \textbf{20.7}
& \cellcolor{blue!8} \textbf{73.6}
& \cellcolor{blue!8} 97.4
& \cellcolor{blue!8} 83.6
& \cellcolor{blue!8} 82.4
& \cellcolor{blue!8} \textbf{68.5}
& \cellcolor{blue!8} \textbf{30.2}
& \cellcolor{blue!8} \textbf{87.6}
& \cellcolor{blue!8} 68.8
& \cellcolor{blue!8} 68.1 \\

& \cellcolor{blue!8}\scriptsize$\Delta$ vs. SFT
& \cellcolor{blue!8}\textcolor{green!50!black}{\scriptsize$\uparrow$4.9}
& \cellcolor{blue!8}\textcolor{green!50!black}{\scriptsize$\uparrow$2.9}
& \cellcolor{blue!8}\textcolor{green!50!black}{\scriptsize$\uparrow$0.6}
& \cellcolor{blue!8}\textcolor{green!50!black}{\scriptsize$\uparrow$0.5}
& \cellcolor{blue!8}\textcolor{green!50!black}{\scriptsize$\uparrow$2.2}
& \cellcolor{blue!8}\textcolor{green!50!black}{\scriptsize$\uparrow$2.1}
& \cellcolor{blue!8}\textcolor{green!50!black}{\scriptsize$\uparrow$\textbf{2.2}}
& \cellcolor{blue!8}\textcolor{green!50!black}{\scriptsize$\uparrow$2.5}
& \cellcolor{blue!8}\textcolor{green!50!black}{\scriptsize$\uparrow$2.2}
& \cellcolor{blue!8}\textcolor{green!50!black}{\scriptsize$\uparrow$0.5}
& \cellcolor{blue!8}\textcolor{green!50!black}{\scriptsize$\uparrow$1.5}
& \cellcolor{blue!8}\textcolor{green!50!black}{\scriptsize$\uparrow$1.5}
& \cellcolor{blue!8}\textcolor{green!50!black}{\scriptsize$\uparrow$2.5}
& \cellcolor{blue!8}\textcolor{green!50!black}{\scriptsize$\uparrow$1.4}
& \cellcolor{blue!8}\textcolor{green!50!black}{\scriptsize$\uparrow$2.5}
& \cellcolor{blue!8}\textcolor{green!50!black}{\scriptsize$\uparrow$3.2}
& \cellcolor{blue!8}\textcolor{green!50!black}{\scriptsize$\uparrow$\textbf{2.0}} \\

\bottomrule
\end{tabular}
}

\end{table*}

\textbf{Beyond Individual Expert Teachers.}
Table~\ref{tab:rl_experts} further shows that the final MMVistaReason models go beyond simply selecting either expert teacher. After multi-teacher integration, the Real-World Avg. reaches 82.0/83.8 at 4B/9B, close to the corresponding experts at 82.7/84.5, while the Analytical Avg. reaches 66.6/68.1, matching or nearly matching the Analytical Experts at 66.6/68.2. This shows that MOPD preserves most specialized gains from both experts within a unified model across scales.

At the benchmark level, MVR inherits improvements from both experts across scales. On real-world tasks, OSWorld-G improves from 54.0 to 58.5 at 4B and from 59.9 to 64.8 at 9B, while CountBenchQA improves from 88.7 to 91.5 and from 90.7 to 92.8, respectively. On analytical tasks, SFE improves from 17.8 to 19.8 and from 18.2 to 20.7, while CharXiv$_{\mathrm{reas.}}$ improves from 63.6 to 66.9 and from 65.6 to 68.8. Notably, MVR improves over the corresponding SFT model on all 15 benchmarks at both scales, indicating broad capability integration rather than optimization toward a single reasoning group.

Notably, MVR surpasses both expert teachers on several benchmarks, including CountBenchQA, SFE, and MMMU$_{\mathrm{val}}$. This suggests that MOPD is not merely a compromise between specialists, but can combine complementary supervision into a unified model that exceeds either expert. Such gains may arise because on-policy distillation exposes each teacher to student-generated states, allowing expert guidance to refine weaknesses of the shared policy. Together, these results support our specialize-then-integrate design, where domain-specific RL develops complementary capabilities and MOPD consolidates them without inheriting cross-domain trade-offs.

\begin{figure*}[t]
    \centering
    \includegraphics[width=1\textwidth]{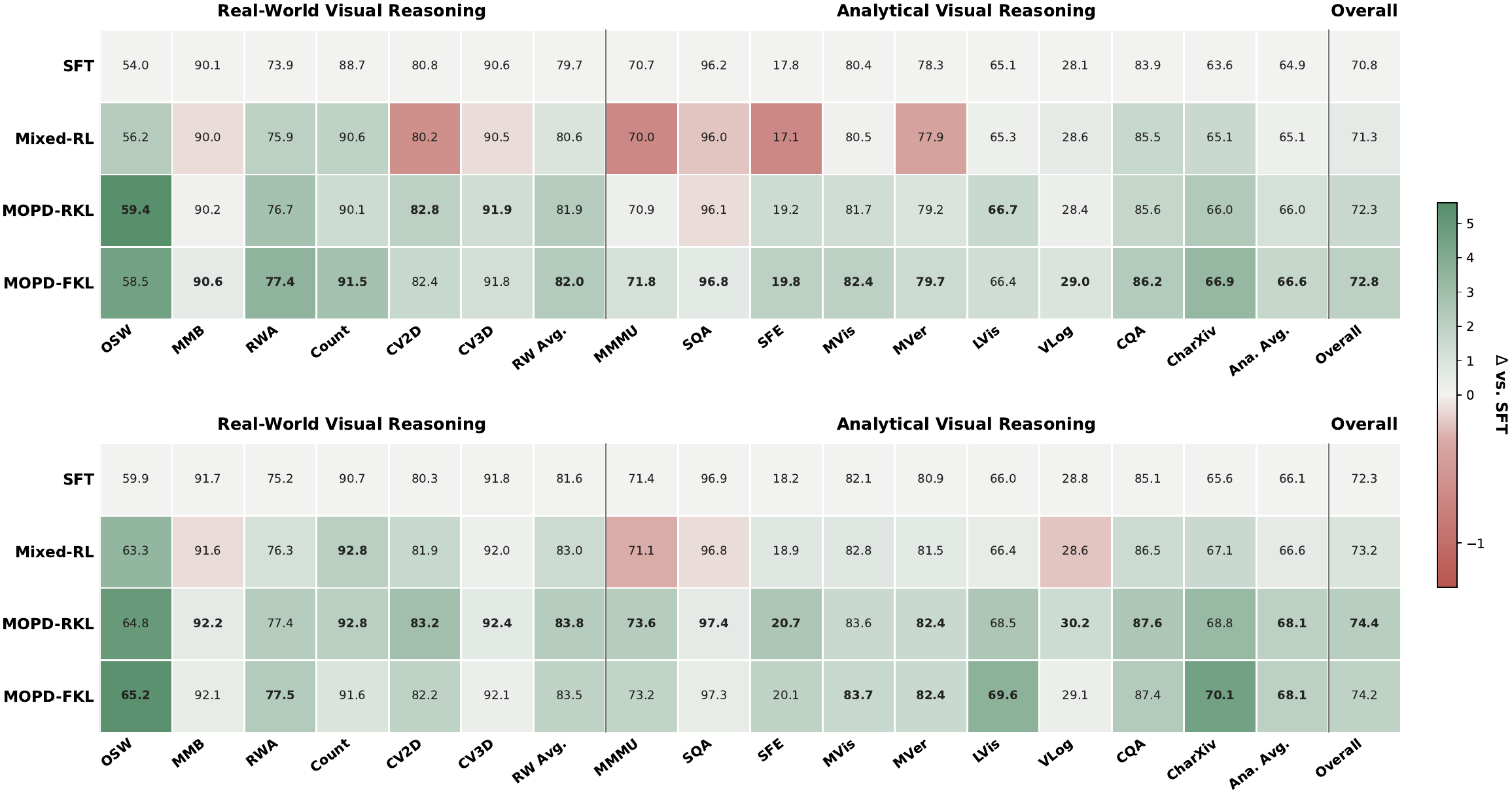}
    \caption{Comparison of Mixed-RL and MOPD under different KL directions across model scales. Colors indicate changes over MVR-SFT for 4B (top) and 9B (bottom) models.}
    \label{fig:mopd_mixrl_heatmap}
\end{figure*}

\textbf{Multi-Teacher Integration Compared with Mixed-Domain RL.}
Figure~\ref{fig:mopd_mixrl_heatmap} compares mixed-domain RL with multi-teacher on-policy distillation across model scales. Mixed-RL improves performance at both scales, raising the Overall Avg. from 70.8 to 71.3 at 4B and from 72.3 to 73.2 at 9B. Real-World and Analytical averages improve from 79.7 to 80.6 and 64.9 to 65.1 at 4B, and from 81.6 to 83.0 and 66.1 to 66.6 at 9B. The larger 9B gains suggest that larger models better absorb heterogeneous RL signals.

However, Mixed-RL still introduces uneven adaptation and local negative transfer. At 4B, it underperforms the SFT baseline on 7 of 15 benchmarks, with the largest drop of 0.7 points on MMMU$_{\mathrm{val}}$ and SFE. At 9B, only four benchmarks regress and the largest drop shrinks to 0.3 points. This suggests that larger model capacity mitigates interference from heterogeneous RL signals, but does not eliminate it, motivating a more reliable strategy for integrating domain-specific capabilities.

In contrast, multi-teacher on-policy distillation provides an integration strategy. At 4B, MOPD-RKL and MOPD-FKL reach 72.3 and 72.8 overall, while at 9B they reach 74.4 and 74.2, outperforming Mixed-RL. The preferred divergence varies with scale: FKL performs better at 4B, whereas RKL is stronger at 9B. Nevertheless, their gap is small compared with the advantage over Mixed-RL. Moreover, the best MOPD variant at each scale improves or matches Mixed-RL on all 15 benchmarks. This suggests that expert-based on-policy integration matters more than the exact KL direction, enabling domain-matched supervision to consolidate complementary capabilities with less interference.

\textbf{Effect of KL Direction in MOPD.}
We further study the effect of KL direction in MOPD across model scales. Both Forward-KL and Reverse-KL outperform the SFT baseline and Mixed-RL in overall performance, confirming the effectiveness of expert distillation on student-induced on-policy trajectories. At 4B, Forward-KL outperforms Reverse-KL (72.8 vs. 72.3), whereas at 9B, Reverse-KL performs slightly better (74.4 vs. 74.2), revealing a scale-dependent preference for KL direction.

Specifically, the two KL directions exhibit task- and scale-dependent preferences. At 4B, Reverse-KL performs better on several real-world benchmarks, while Forward-KL is stronger on most remaining tasks. At 9B, Reverse-KL becomes more competitive across benchmarks. This pattern suggests that its mode-seeking behavior may become more beneficial as student capacity increases, whereas smaller students benefit more from the broader distribution matching of Forward-KL.

Since both RL experts are initialized from the same SFT generalist, their output distributions remain relatively aligned with the student, allowing both KL objectives to provide effective supervision. Forward-KL offers broader distribution matching that may be easier for smaller students to absorb, whereas Reverse-KL can better exploit sharper expert preferences at larger scale. Importantly, the gap between the two objectives remains small at both scales, suggesting that MOPD is robust to the exact KL direction and that expert-based on-policy integration is the primary source of improvement.

\section{Conclusion}

In this work, we present \textbf{MMVistaReason}, an open-data framework and adaptive training recipe for reliable multimodal reasoning. Rather than treating multimodal reasoning as a problem driven only by model scaling, we view it as a joint challenge of data construction, difficulty-aware sample selection, expert specialization, and capability integration. Through systematic data cleaning, answer normalization, model-aware difficulty filtering, cascaded rejection sampling, and Answer-Likelihood Best-of-N trajectory selection, MMVistaReason transforms heterogeneous open data into high-quality and verifiable reasoning supervision. Building on this corpus, we develop a training pipeline that establishes strong SFT generalists, then trains complementary RL experts for reasoning scenarios, and finally integrates their specialized capabilities through multi-teacher on-policy distillation. Extensive experiments across 4B and 9B scales show that MMVistaReason achieves strong and balanced performance across diverse multimodal reasoning benchmarks, while analyses reveal that supervision difficulty, trajectory quality, and integration behavior interact with student model capacity. In particular, domain-specific specialization followed by MOPD better consolidates complementary expert capabilities with reduced cross-domain interference. Overall, MMVistaReason provides a practical and scalable path for building reliable multimodal reasoning models from open data.

\clearpage
\newpage
\bibliographystyle{plainnat}
\setcitestyle{numbers}
\bibliography{paper}

\clearpage

\newpage
\beginappendix

\section{Additional Data Analysis}

\subsection{SFT Data Statistics}

We provide additional analysis of the SFT corpus (MMVistaReason-SFT-528K), including its open-data sources, subset composition, difficulty characteristics, and task diversity.

\textbf{Open-Data Source Composition.}
We further analyze the source composition of MMVistaReason-SFT-528K. As shown in Table~\ref{tab:sft_source_statistics}, the corpus is constructed from a diverse collection of open multimodal datasets spanning science, mathematics, spatial reasoning, charts, visual logic, and general visual understanding. Compared with existing multimodal reasoning recipes, our corpus draws from a broader range of sources and task scenarios. The broad source coverage highlights the heterogeneous open-data foundation of MMVistaReason and supports supervision across reasoning scenarios.

\begin{table}[!htbp]
\centering
\caption{
Detailed source composition of \textbf{MMVistaReason-SFT-528K}.
We report the number of samples and response tokens for each source subset.
}
\label{tab:sft_source_statistics}

\footnotesize
\setlength{\tabcolsep}{8pt}
\renewcommand{\arraystretch}{1.05}

\begin{tabular}{lrr@{\hspace{18pt}}lrr}
\toprule
\textbf{Subset Name} & \textbf{Samples} & \textbf{Tokens} &
\textbf{Subset Name} & \textbf{Samples} & \textbf{Tokens} \\
\midrule

SciMM \cite{sciverse_opendatalab}
& 247,292 & 291,185,096
& vero\_captioning\_IF \cite{sarch2026vero}
& 4,506 & 2,755,634 \\

MMR1 \cite{leng2025mmr1}
& 77,116 & 322,402,209
& PRISM \cite{wang2026beyond}
& 3,831 & 11,606,008 \\

vero\_spatial\_action \cite{sarch2026vero}
& 30,153 & 48,377,896
& Euclid30K \cite{euclid}
& 3,038 & 14,260,100 \\

FineVision \cite{finevision}
& 22,799 & 42,814,429
& WaltonColdStart \cite{walton}
& 2,932 & 7,034,074 \\

vero\_chart\_ocr \cite{sarch2026vero}
& 21,998 & 19,464,798
& ViRL39K \cite{vl-rethinker}
& 2,236 & 6,383,976 \\

deepvision\_math \cite{sun2026deepvision}
& 19,550 & 88,001,303
& LLaVA-CoT \cite{xu2025llava}
& 2,018 & 906,254 \\

vero\_knowledge\_recognition \cite{sarch2026vero}
& 19,058 & 9,900,384
& MMK12 \cite{mmk12}
& 1,215 & 5,194,661 \\

GameQA-140K \cite{gameqa}
& 16,933 & 45,353,567
& Zebra-CoT-Physics \cite{zebracot}
& 874 & 2,044,486 \\

vero\_stem \cite{sarch2026vero}
& 16,333 & 48,929,432
& WeMath2 \cite{qiao2026we}
& 841 & 2,889,230 \\

vero\_counting\_grounding\_search \cite{sarch2026vero}
& 13,173 & 9,424,096
& mmopenr1-8k \cite{mmopenr1}
& 690 & 1,799,171 \\

BMMR \cite{bmmr}
& 10,504 & 32,139,488
& VisualSphinx \cite{visualsphinx}
& 617 & 1,682,056 \\

deepvision\_visual\_logic \cite{sun2026deepvision}
& 9,709 & 28,925,351
& PuzzleQA \cite{puzzlevqa}
& 393 & 578,248 \\

\midrule
\multicolumn{6}{c}{
\textbf{Total Samples: 527,809}
\qquad
\textbf{Total Tokens: 1,044,051,947}
} \\
\bottomrule
\end{tabular}
\end{table}

\textbf{Difficulty Characteristics of SFT Subsets} To further characterize the difficulty differences among SFT subsets produced by the cascaded teacher distillation process, we analyze their SFT rollout \textbf{pass-rate} distributions. As shown in Table~\ref{tab:sft_difficulty}, subsets distilled from stronger teachers exhibit lower average pass rates, indicating progressively harder supervision signals.

\begin{table}[!htbp]
\centering
\caption{Difficulty distribution of different SFT subsets generated by cascaded teacher distillation.}
\label{tab:sft_difficulty}
\resizebox{\columnwidth}{!}{
\begin{tabular}{lccccc}
\toprule
\textbf{SFT Subset} &
\textbf{Mastery=0.25} &
\textbf{Mastery=0.50} &
\textbf{Mastery=0.75} &
\textbf{Mastery=1.00} &
\textbf{Weighted Mean} \\
\midrule
MVR-SFT-49K (122B)  & 69.26\% & 19.53\% & 7.61\%  & 3.61\%  & 0.3639 \\
MVR-SFT-124K (27B) & 49.74\% & 25.50\% & 15.17\% & 9.59\%  & 0.4616 \\
MVR-SFT-355K (9B)  & 40.49\% & 26.23\% & 19.06\% & 14.23\% & 0.5176 \\
\bottomrule
\end{tabular}
}
\end{table}

\begin{table*}[t]
\centering
\caption{Question subtype distribution of MVR-SFT-528K constructed by different teacher models. Each entry reports the number of samples and the corresponding ratio within each subset.}
\scriptsize
\label{tab:sft_subtype_distribution}
\resizebox{\textwidth}{!}{
\begin{tabular}{lcccc}
\toprule
\textbf{Subtype} &
\textbf{122B Teacher} &
\textbf{27B Teacher} &
\textbf{9B Teacher} &
\textbf{Overall} \\
\midrule
Numeric 
& 12,058 (24.65\%) 
& 37,806 (30.49\%) 
& 110,705 (31.19\%) 
& 160,569 (30.42\%) \\

Short Phrase 
& 16,985 (34.72\%) 
& 41,626 (33.57\%) 
& 89,283 (25.16\%) 
& 147,894 (28.02\%) \\

Multiple Choice 
& 4,665 (9.54\%) 
& 19,954 (16.09\%) 
& 72,450 (20.42\%) 
& 97,069 (18.39\%) \\

Entity Name 
& 7,188 (14.70\%) 
& 16,611 (13.39\%) 
& 51,856 (14.61\%) 
& 75,655 (14.33\%) \\

Yes/No 
& 748 (1.53\%) 
& 2,562 (2.07\%) 
& 13,504 (3.81\%) 
& 16,814 (3.19\%) \\

Position 
& 6,358 (13.00\%) 
& 2,720 (2.19\%) 
& 5,045 (1.42\%) 
& 14,123 (2.68\%) \\

Counting 
& 769 (1.57\%) 
& 2,273 (1.83\%) 
& 10,429 (2.94\%) 
& 13,471 (2.55\%) \\

Color 
& 90 (0.18\%) 
& 305 (0.25\%) 
& 1,055 (0.30\%) 
& 1,450 (0.27\%) \\

Category 
& 43 (0.09\%) 
& 135 (0.11\%) 
& 529 (0.15\%) 
& 707 (0.13\%) \\

Short Fact 
& 9 (0.02\%) 
& 18 (0.01\%) 
& 30 (0.01\%) 
& 57 (0.01\%) \\

\midrule
\textbf{Total}
& \textbf{48,913 (100\%)}
& \textbf{124,010 (100\%)}
& \textbf{354,886 (100\%)}
& \textbf{527,809 (100\%)}\\
\bottomrule
\end{tabular}
}
\end{table*}

\textbf{Question Type Distribution of SFT Subsets} We further analyze the question type distribution of different SFT subsets. As shown in Table~\ref{tab:sft_subtype_distribution}, the constructed corpus covers diverse multimodal reasoning formats, including numeric, short-answer, multiple-choice, and entity-based questions. The complementary distributions across teacher-derived subsets demonstrate the diversity of supervision introduced by the cascaded distillation strategy.

\textbf{Domain-wise Token Statistics}
We further report detailed response-token statistics for each reasoning domain in Table~\ref{tab:sft_domain_token_stats}. The domains exhibit substantial differences in both data scale and reasoning length. Mathematics accounts for only 24.76\% of the samples but contributes 49.70\% of the response tokens, with the highest mean length of 3,970.5 tokens and a P95 length of 12,188 tokens. Logic/Game/Puzzle is similarly reasoning-intensive, contributing a larger proportion of tokens than samples. Spatial has a moderate median length but a pronounced long-tailed distribution, with its P95 reaching 6,344 tokens. In comparison, Science and Chart/Table/Doc contain large numbers of samples with relatively moderate response lengths, whereas General and GUI/Web are dominated by short-form responses. These statistics demonstrate that MVR-SFT balances large-scale visual supervision with computationally intensive long-form reasoning data of varying complexity.

  \begin{table*}[t]
  \centering
  \fontsize{9.5pt}{10.5pt}\selectfont
  \begin{tabular}{lrrrrrrrr}
  \toprule
  \textbf{Domain} &
  \textbf{\#Samples} &
  \textbf{Tokens} &
  \textbf{Mean} &
  \textbf{Std.} &
  \textbf{Median} &
  \textbf{P75} &
  \textbf{P95} &
  \textbf{Max} \\
  \midrule

  Science
  & 175,007
  & 208.90M
  & 1,193.7
  & 1,365.4
  & 681
  & 1,420
  & 3,723
  & 16,369 \\

  Mathematics
  & 130,677
  & 518.85M
  & 3,970.5
  & 3,739.1
  & 2,669
  & 5,875
  & 12,188
  & 16,381 \\

  Logic/Game/Puzzle
  & 47,140
  & 131.24M
  & 2,784.1
  & 2,448.5
  & 2,107
  & 3,919
  & 7,421
  & 16,375 \\

  Chart/Table/Doc
  & 109,250
  & 121.95M
  & 1,116.3
  & 1,314.4
  & 588
  & 1,393
  & 3,708
  & 16,039 \\

  General
  & 32,165
  & 15.75M
  & 489.5
  & 635.3
  & 277
  & 513
  & 1,532
  & 13,725 \\

  Spatial
  & 23,470
  & 42.31M
  & 1,802.6
  & 2,182.1
  & 978
  & 2,238
  & 6,344
  & 16,374 \\

  GUI/Web
  & 10,100
  & 5.06M
  & 500.5
  & 878.3
  & 255
  & 426
  & 1,689
  & 16,192 \\

  \midrule
  \textbf{All Domains}
  & \textbf{527,809}
  & \textbf{1.04B}
  & \textbf{1,978.1}
  & \textbf{2,607.8}
  & \textbf{906}
  & \textbf{2,359}
  & \textbf{7,625}
  & \textbf{16,381} \\

  \bottomrule
  \end{tabular}

\caption{Domain-wise response-token statistics of MVR-SFT-528K.
  Token statistics are computed over model responses using the Qwen3.5 tokenizer for all seven reasoning domains.}

  \label{tab:sft_domain_token_stats}
  \end{table*}

\subsection{RL Data Statistics}

We provide a comprehensive statistical analysis of the RL data used to construct the domain-specific experts. This analysis offers a systematic overview of the data composition and characteristics.
  
\textbf{Difficulty Distribution of RL Experts.}
We further compare the difficulty characteristics of RL expert training sets across different model scales. As shown in Table~\ref{tab:rl_difficulty_distribution}, both Analytical and Real-World RL subsets concentrate around the medium difficulty range, with a median SFT rollout pass rate of 0.5. Analytical RL is consistently more challenging, with mean pass rates of 0.4502 and 0.4503 at 4B and 9B, respectively, compared with 0.4981 for Real-World RL at both scales. Notably, the overall difficulty profiles remain highly consistent across model scales, showing that our scale-specific frontier selection constructs RL subsets with stable and well-matched difficulty distributions.

\begin{table}[!htbp]
  \centering
  \caption{Difficulty distribution of 4B and 9B RL experts measured by SFT rollout pass rate.}
  \label{tab:rl_difficulty_distribution}
  \small
  \begin{tabular}{lccccccccc}
  \toprule
  \textbf{RL Expert} &
  \textbf{0.125} &
  \textbf{0.250} &
  \textbf{0.375} &
  \textbf{0.500} &
  \textbf{0.625} &
  \textbf{0.750} &
  \textbf{0.875} &
  \textbf{Mean} &
  \textbf{Median} \\
  \midrule
  4B Analytical RL
  & 44 & 1,955 & 6,654 & 7,633 & 2,620 & 194 & 25
  & 0.4502 & 0.500 \\

  4B Real-World RL
  & 0 & 1,348 & 2,939 & 3,270 & 2,624 & 1,416 & 0
  & 0.4981 & 0.500 \\

  9B Analytical RL
  & 47 & 2,085 & 7,097 & 8,142 & 2,795 & 207 & 27
  & 0.4503 & 0.500 \\

  9B Real-World RL
  & 0 & 1,393 & 3,036 & 3,377 & 2,711 & 1,463 & 0
  & 0.4981 & 0.500 \\
  \midrule
  Overall
  & 91 & 6,781 & 19,726 & 22,422 & 10,750 & 3,280 & 52
  & 0.4681 & 0.500 \\
  \bottomrule
  \end{tabular}
  \end{table}

\textbf{Question Type Distribution of RL Experts.}
We further analyze the task-type distribution of RL experts across model scales. As shown in Table~\ref{tab:rl_question_type}, Analytical RL is dominated by multiple-choice and numeric reasoning tasks, while Real-World RL contains substantially more counting and position-related questions. These domain-specific patterns remain highly consistent between 4B and 9B, indicating stable and well-preserved task compositions across scales. Moreover, these task types generally have well-defined answer spaces, enabling reliable automatic verification and reducing reward ambiguity during reinforcement learning. The complementary yet stable distributions provide diverse and verifiable supervision signals for effective domain-specific RL training.

 \begin{table}[!htbp]
  \centering
  \caption{Question type distribution of 4B and 9B RL experts.}
  \label{tab:rl_question_type}
  \small
  \begin{tabular}{lcccc}
  \toprule
  \textbf{Question Type} &
  \textbf{4B Analytical} &
  \textbf{4B Real-World} &
  \textbf{9B Analytical} &
  \textbf{9B Real-World} \\
  \midrule

  Multiple Choice
  & 11,269 (58.92\%)
  & 4,975 (42.90\%)
  & 11,669 (57.20\%)
  & 5,002 (41.75\%) \\

  Numeric
  & 5,250 (27.45\%)
  & 254 (2.19\%)
  & 5,007 (24.54\%)
  & 287 (2.40\%) \\

  Counting
  & 396 (2.07\%)
  & 3,404 (29.35\%)
  & 446 (2.19\%)
  & 3,482 (29.07\%) \\

  Position
  & 22 (0.12\%)
  & 1,943 (16.75\%)
  & 37 (0.18\%)
  & 2,052 (17.13\%) \\

  Short Phrase
  & 1,107 (5.79\%)
  & 560 (4.83\%)
  & 1,813 (8.89\%)
  & 659 (5.50\%) \\

  Entity Name
  & 778 (4.07\%)
  & 175 (1.51\%)
  & 978 (4.79\%)
  & 205 (1.71\%) \\

  Yes/No
  & 248 (1.30\%)
  & 221 (1.91\%)
  & 360 (1.76\%)
  & 242 (2.02\%) \\

  Color
  & 46 (0.24\%)
  & 58 (0.50\%)
  & 66 (0.32\%)
  & 49 (0.41\%) \\

  Category
  & 9 (0.05\%)
  & 6 (0.05\%)
  & 23 (0.11\%)
  & 2 (0.02\%) \\

  Short Fact
  & 0 (0.00\%)
  & 1 (0.01\%)
  & 1 (0.00\%)
  & 0 (0.00\%) \\

  \midrule
  \textbf{Total}
  & \textbf{19,125 (100\%)}
  & \textbf{11,597 (100\%)}
  & \textbf{20,400 (100\%)}
  & \textbf{11,980 (100\%)} \\

  \bottomrule
  \end{tabular}
  \end{table}

\section{Experimental Details}

\subsection{Distillation Details}
For cascaded reasoning trajectory distillation, we employ Qwen3.5-9B, Qwen3.5-27B, and Qwen3.5-122B-A10B as teacher models, deployed on 1, 2, and 4 NVIDIA A100 GPUs with 80\,GB memory, respectively. To control the computational cost of large-scale trajectory generation, all teacher models are run in instruct mode (\texttt{enable\_thinking=False}). For each input sample, we generate four candidate trajectories and set the maximum generation length to 16,384 tokens. Under this deployment configuration, the 9B, 27B, and 122B-A10B teachers process approximately 200, 120, and 80 samples per hour, respectively, corresponding to roughly 800, 480, and 320 generated trajectories per hour. This cascaded design reduces distillation cost by routing only difficult samples to larger teachers.

\begin{table}[htbp]
\centering
\caption{Training hyperparameters for SFT, RL, and MOPD.}
\label{tab:training_hyperparams}
\setlength{\tabcolsep}{4pt}
\renewcommand{\arraystretch}{1.08}
\small

\begin{minipage}[t]{0.32\textwidth}
\centering
\textbf{SFT Params.}\\[1mm]
\begin{tabular}{lc}
\toprule
\textbf{Parameter} & \textbf{Value} \\
\midrule
Framework & LLaMA-Factory \\
Fine-tuning type & Full fine-tuning \\
GPU Number & 8 \\
DeepSpeed & ZeRO-2 \\
Flash attention & FlashAttention-2 \\
Liger kernel & Enabled \\
Packing & Enabled \\
Cutoff length & 16,384 \\
Image min pixels & 1,024 \\
Image max pixels & 589,824 \\
Global batch size & 32 \\
Learning rate & $1.0\times10^{-5}$ \\
Scheduler & Cosine \\
Warmup ratio & 0.1 \\
Training epochs & 2 \\
\bottomrule
\end{tabular}
\end{minipage}
\hfill
\begin{minipage}[t]{0.32\textwidth}
\centering
\textbf{RL Params.}\\[1mm]
\begin{tabular}{lc}
\toprule
\textbf{Parameter} & \textbf{Value} \\
\midrule
Framework & VeRL \\
RL objective & GSPO \\
GPU Number & 8 \\
Global batch size & 256 \\
Max prompt length & 4,096 \\
Max response length & 16,384 \\
Rollouts & 8 \\
Temperature & 1.0 \\
Top-$p$ & 0.95 \\
Actor learning rate & $5\times10^{-7}$ \\
Weight decay & 0.1 \\
Warmup steps & 20 \\
Training steps & 200 \\
Judge temperature & 0.1 \\
Judge top-$p$ & 1.0 \\
\bottomrule
\end{tabular}
\end{minipage}
\hfill
\begin{minipage}[t]{0.32\textwidth}
\centering
\textbf{MOPD Params.}\\[1mm]
\begin{tabular}{lc}
\toprule
\textbf{Parameter} & \textbf{Value} \\
\midrule
Framework & VeRL \\
Student GPUs & 4 \\
GPUs per teacher & 2 \\
Global batch size & 256 \\
Max prompt length & 4,096 \\
Max response length & 16,384 \\
Rollouts per prompt & 2 \\
Temperature & 1.0 \\
Top-$p$ & 0.95 \\
Actor learning rate & $5.0\times10^{-7}$ \\
Weight decay & 0.1 \\
Warmup steps & 20 \\
Training steps & 300 \\
Top-$K$ & 32 \\
LR schedule & Constant \\
\bottomrule
\end{tabular}
\end{minipage}

\end{table}

\subsection{Training Details}

\textbf{Supervised Fine-Tuning.}
We conduct SFT with LLaMA-Factory on 8 NVIDIA A100 GPUs with 80GB memory each. We use full-parameter fine-tuning with packed sequences, FlashAttention-2, and a cutoff length of 16,384 tokens for efficient long-context training. The global batch size is set to 32. Other SFT hyperparameters are summarized in Table~\ref{tab:training_hyperparams}.

\textbf{Reinforcement Learning.}
We train RL experts with VeRL on 8 NVIDIA H200 GPUs with 141GB memory each. We adopt GSPO as the RL objective and sample 8 rollouts per prompt. The maximum prompt and response lengths are set to 4,096 and 16,384, respectively. We use an answer-only judge reward during RL training. Detailed RL hyperparameters are shown in Table~\ref{tab:training_hyperparams}.

\textbf{Multi-Teacher On-Policy Distillation.}
We perform MOPD with VeRL on NVIDIA H200 GPUs with 141GB memory. The student uses 4 GPUs, while each teacher uses 2 GPUs. The student generates on-policy responses, and the corresponding teacher provides top-$K$ token-level supervision with $K=32$. Other MOPD settings are reported in Table~\ref{tab:training_hyperparams}.

\subsection{Evaluation Details}

We conduct evaluation using VLMEvalKit~\cite{duan2024vlmevalkit}. All evaluations are performed on NVIDIA H200 GPUs with 141GB memory. For evaluation metrics, we replace traditional exact string matching with CompassVerifier~\cite{liu2025compassverifier}, which adopts an LLM-as-a-Judge to more reliably assess response correctness. 
Following OpenDataArena~\cite{cai2025opendataarena} and the official evaluation setting, we use deterministic decoding with temperature 0.0, top-$p$ 1.0, and top-$k$ -1. 
We further apply a repetition penalty of 1.05 and set the maximum response length to 16,384 tokens. Responses that do not produce a valid final answer within this budget are not counted as correct. For benchmarks with unusually large performance deviations, we additionally conduct manual inspections to verify the reliability of the evaluation results.

\section{Prompt Templates}

In this section, we present the prompts used in our experiments.

\subsection{MLLM-Assisted Cleaning and Annotation}

\refstepcounter{prompt}\label{pmt:language_standardization}
\begin{tcolorbox}[
  colback=gray!5,
  colframe=black!75,
  width=\columnwidth,
  title={Prompt \theprompt: Language Standardization},
  breakable,
  boxrule=0.8pt,
  arc=2pt,
  left=6pt,right=6pt,top=6pt,bottom=6pt
]
\small
\raggedright
\sloppy

\textbf{Role.} You are a careful multilingual data curator for multimodal visual reasoning tasks.

\medskip
\textbf{Inputs.} An image and a raw question that may contain non-English text, mathematical expressions, variables, units, option labels, or visual identifiers.

\medskip
\textbf{Task.} Standardize the question into English while preserving the original task meaning. Follow these rules:
\begin{itemize}[leftmargin=*, itemsep=1pt, topsep=1pt]
  \item If the question is not in English, translate it into fluent and natural English.
  \item If the question is already in English, keep its meaning unchanged and only fix minor language issues when necessary.
  \item Preserve all mathematical formulas, variables, numbers, units, option labels, and visual identifiers.
  \item Do not change the answer space, add new constraints, remove options, or simplify the reasoning required by the question.
  \item Do not solve the problem or provide the final answer.
\end{itemize}

\medskip
\textbf{Output (strict).} The last line must be a single JSON object with exactly these keys:
\begin{itemize}[leftmargin=*, itemsep=1pt, topsep=1pt]
  \item \texttt{standardized\_question}, \texttt{is\_translated}, \texttt{reason}
\end{itemize}
No extra keys. No fences. No text after the JSON.

\medskip
\textbf{JSON schema example.}
\begin{quote}\ttfamily\footnotesize\raggedright\sloppy
\{ "standardized\_question":"", "is\_translated":false, "reason":"" \}
\end{quote}

\textbf{Image.}
\begin{quote}\ttfamily
<IMAGE\_START>\\
\{image\}\\
<IMAGE\_END>
\end{quote}

\textbf{Raw Question.}
\begin{quote}\ttfamily
<QUESTION\_START>\\
\{question\}\\
<QUESTION\_END>
\end{quote}

\end{tcolorbox}

\refstepcounter{prompt}\label{pmt:irrelevant_content_removal}
\begin{tcolorbox}[
  colback=gray!5,
  colframe=black!75,
  width=\columnwidth,
  title={Prompt \theprompt: Irrelevant Content Removal},
  breakable,
  boxrule=0.8pt,
  arc=2pt,
  left=6pt,right=6pt,top=6pt,bottom=6pt
]
\small
\raggedright
\sloppy

\textbf{Role.} You are a strict data curator for multimodal visual reasoning tasks.

\medskip
\textbf{Inputs.} An image and a standardized question that may contain unrelated or distracting content.

\medskip
\textbf{Task.} Remove irrelevant content from the question while preserving the original task meaning. Follow these rules:
\begin{itemize}[leftmargin=*, itemsep=1pt, topsep=1pt]
  \item Remove unrelated links, advertisements, watermarks, source notes, problem indices, score annotations, and irrelevant metadata.
  \item Remove boilerplate instructions that are not necessary for answering the visual reasoning question.
  \item Preserve all task-relevant information, including formulas, variables, numbers, units, option labels, visual identifiers, and constraints.
  \item Do not remove content that is necessary to understand or answer the question.
  \item Do not solve the problem or provide the final answer.
\end{itemize}

\medskip
\textbf{Output (strict).} The last line must be a single JSON object with exactly these keys:
\begin{itemize}[leftmargin=*, itemsep=1pt, topsep=1pt]
  \item \texttt{error type}, \texttt{corrected text}, \texttt{reason}
\end{itemize}
The value of \texttt{error type} should be a list. If no irrelevant content is found, return an empty list and keep the original question as \texttt{corrected text}. No extra keys. No fences. No text after the JSON.

\medskip
\textbf{JSON schema example.}
\begin{quote}\ttfamily\footnotesize\raggedright\sloppy
\{ "error type":[], "corrected text":"", "reason":"" \}
\end{quote}

\medskip
\textbf{Example.}
If the standardized question is ``What is this? https://example.com'', remove the unrelated link:
\begin{quote}\ttfamily\footnotesize\raggedright\sloppy
\{ "error type":["irrelevant content"], "corrected text":"What is this?", "reason":"The URL is unrelated to the visual reasoning question and is removed." \}
\end{quote}

\textbf{Image.}
\begin{quote}\ttfamily
<IMAGE\_START>\\
\{image\}\\
<IMAGE\_END>
\end{quote}

\textbf{Standardized Question.}
\begin{quote}\ttfamily
<QUESTION\_START>\\
\{question\}\\
<QUESTION\_END>
\end{quote}

\end{tcolorbox}

\refstepcounter{prompt}\label{pmt:answerability_filtering}
\begin{tcolorbox}[
  colback=gray!5,
  colframe=black!75,
  width=\columnwidth,
  title={Prompt \theprompt: Answerability Filtering},
  breakable,
  boxrule=0.8pt,
  arc=2pt,
  left=6pt,right=6pt,top=6pt,bottom=6pt
]
\small
\raggedright
\sloppy

\textbf{Role.} You are a strict data curator for multimodal visual reasoning tasks.

\medskip
\textbf{Inputs.} An image, a cleaned question, and a reference answer.

\medskip
\textbf{Task.} Determine whether the sample can be reliably answered from the given multimodal input. Follow these rules:
\begin{itemize}[leftmargin=*, itemsep=1pt, topsep=1pt]
  \item Mark the sample as answerable only if the question can be answered from the image and the provided text.
  \item Mark the sample as unanswerable if the question is incomplete, ambiguous, missing key context, or depends on unavailable information.
  \item Mark the sample as unanswerable if it requires external actions, web browsing, file access, interaction with an interface, or information beyond the given image and text.
  \item Mark the sample as unanswerable if it is not a real question, such as a captioning-only request, open-ended discussion, subjective preference, or instruction without a verifiable answer.
  \item Do not solve the problem or modify the question.
\end{itemize}

\medskip
\textbf{Output (strict).} The last line must be a single JSON object with exactly these keys:
\begin{itemize}[leftmargin=*, itemsep=1pt, topsep=1pt]
  \item \texttt{answerable}, \texttt{reason}
\end{itemize}
No extra keys. No fences. No text after the JSON.

\medskip
\textbf{JSON schema example.}
\begin{quote}\ttfamily\footnotesize\raggedright\sloppy
\{ "answerable":false, "reason":"" \}
\end{quote}

\medskip
\textbf{Example.}
If the cleaned question is ``Open the website shown in the image and report the current price'', the sample should be marked as unanswerable because it requires external web access:
\begin{quote}\ttfamily\footnotesize\raggedright\sloppy
\{ "answerable":false, "reason":"The question requires external web browsing and cannot be answered from the given image and text alone." \}
\end{quote}

\textbf{Image.}
\begin{quote}\ttfamily
<IMAGE\_START>\\
\{image\}\\
<IMAGE\_END>
\end{quote}

\textbf{Cleaned Question.}
\begin{quote}\ttfamily
<QUESTION\_START>\\
\{question\}\\
<QUESTION\_END>
\end{quote}

\textbf{Reference Answer.}
\begin{quote}\ttfamily
<ANSWER\_START>\\
\{answer\}\\
<ANSWER\_END>
\end{quote}

\end{tcolorbox}

\refstepcounter{prompt}\label{pmt:instruction_refinement}
\begin{tcolorbox}[
  colback=gray!5,
  colframe=black!75,
  width=\columnwidth,
  title={Prompt \theprompt: Instruction Refinement},
  breakable,
  boxrule=0.8pt,
  arc=2pt,
  left=6pt,right=6pt,top=6pt,bottom=6pt
]
\small
\raggedright
\sloppy

\textbf{Role.} You are a careful data curator for multimodal visual reasoning tasks.

\medskip
\textbf{Inputs.} An image and a cleaned question that may contain low-quality meta-instructions.

\medskip
\textbf{Task.} Refine the question instruction while preserving the original task semantics. Follow these rules:
\begin{itemize}[leftmargin=*, itemsep=1pt, topsep=1pt]
  \item If the question contains instructions that may reduce reasoning quality, rewrite them into instructions that encourage thoughtful reasoning.
  \item Examples of low-quality instructions include ``just give the answer'', ``do not think'', ``give me the final answer only'', or similar phrases.
  \item Replace such instructions with reasoning-friendly guidance, such as ``provide a clear reasoning process before the final answer''.
  \item Preserve the original visual reasoning task, answer space, formulas, variables, numbers, units, option labels, and constraints.
  \item Do not add new task requirements, change the question meaning, remove necessary information, or solve the problem.
\end{itemize}

\medskip
\textbf{Output (strict).} The last line must be a single JSON object with exactly these keys:
\begin{itemize}[leftmargin=*, itemsep=1pt, topsep=1pt]
  \item \texttt{refined\_question}, \texttt{has\_low\_quality\_instruction}, \texttt{rewritten\_instruction}, \texttt{reason}
\end{itemize}
No extra keys. No fences. No text after the JSON.

\medskip
\textbf{JSON schema example.}
\begin{quote}\ttfamily\footnotesize\raggedright\sloppy
\{ "refined\_question":"", "has\_low\_quality\_instruction":false, "rewritten\_instruction":"", "reason":"" \}
\end{quote}

\medskip
\textbf{Example.}
If the cleaned question is ``Please directly answer the question: what are the roots of this equation?'', the low-quality meta-instruction should be rewritten to encourage reasoning:
\begin{quote}\ttfamily\footnotesize\raggedright\sloppy
\{ "refined\_question":"Please provide a clear reasoning process before giving the roots of this equation.", "has\_low\_quality\_instruction":true, "rewritten\_instruction":"Please directly answer the question", "reason":"The original instruction discourages reasoning, so it is replaced with reasoning-friendly guidance while preserving the task meaning." \}
\end{quote}

\textbf{Image.}
\begin{quote}\ttfamily
<IMAGE\_START>\\
\{image\}\\
<IMAGE\_END>
\end{quote}

\textbf{Cleaned Question.}
\begin{quote}\ttfamily
<QUESTION\_START>\\
\{question\}\\
<QUESTION\_END>
\end{quote}

\end{tcolorbox}

\refstepcounter{prompt}\label{pmt:answer_extraction}
\begin{tcolorbox}[
  colback=gray!5,
  colframe=black!75,
  width=\columnwidth,
  title={Prompt \theprompt: Answer Extraction},
  breakable,
  boxrule=0.8pt,
  arc=2pt,
  left=6pt,right=6pt,top=6pt,bottom=6pt
]
\small
\raggedright
\sloppy

\textbf{Role.} You are a careful data curator for multimodal visual reasoning tasks.

\medskip
\textbf{Inputs.} An image, a refined question, and a raw answer that may contain reasoning steps, explanations, options, or noisy text.

\medskip
\textbf{Task.} Extract the final answer from the raw answer field while preserving the original answer meaning. Follow these rules:
\begin{itemize}[leftmargin=*, itemsep=1pt, topsep=1pt]
  \item If the raw answer contains reasoning, explanations, or intermediate steps, extract only the final answer.
  \item If the raw answer contains multiple candidate answers, extract the one explicitly indicated as the final answer.
  \item Preserve important answer details, including option labels, numbers, units, variables, signs, and mathematical expressions.
  \item Do not infer a new answer from the image or question; only extract the answer already present in the raw answer field.
  \item Do not rewrite the question, solve the problem, or add new reasoning.
\end{itemize}

\medskip
\textbf{Output (strict).} The last line must be a single JSON object with exactly these keys:
\begin{itemize}[leftmargin=*, itemsep=1pt, topsep=1pt]
  \item \texttt{extracted\_answer}, \texttt{has\_reasoning\_or\_noise}, \texttt{reason}
\end{itemize}
No extra keys. No fences. No text after the JSON.

\medskip
\textbf{JSON schema example.}
\begin{quote}\ttfamily\footnotesize\raggedright\sloppy
\{ "extracted\_answer":"", "has\_reasoning\_or\_noise":false, "reason":"" \}
\end{quote}

\textbf{Image.}
\begin{quote}\ttfamily
<IMAGE\_START>\\
\{image\}\\
<IMAGE\_END>
\end{quote}

\textbf{Refined Question.}
\begin{quote}\ttfamily
<QUESTION\_START>\\
\{question\}\\
<QUESTION\_END>
\end{quote}

\textbf{Raw Answer.}
\begin{quote}\ttfamily
<ANSWER\_START>\\
\{answer\}\\
<ANSWER\_END>
\end{quote}

\end{tcolorbox}

\refstepcounter{prompt}\label{pmt:answer_canonicalization}
\begin{tcolorbox}[
  colback=gray!5,
  colframe=black!75,
  width=\columnwidth,
  title={Prompt \theprompt: Answer Canonicalization},
  breakable,
  boxrule=0.8pt,
  arc=2pt,
  left=6pt,right=6pt,top=6pt,bottom=6pt
]
\small
\raggedright
\sloppy

\textbf{Role.} You are a careful data curator for multimodal visual reasoning tasks.

\medskip
\textbf{Inputs.} An image, a refined question, and an extracted answer.

\medskip
\textbf{Task.} Normalize the extracted answer into a concise, consistent, and verifiable form while preserving its original meaning. Follow these rules:
\begin{itemize}[leftmargin=*, itemsep=1pt, topsep=1pt]
  \item If the question is multiple-choice and the answer contains option content, recover the corresponding option label from the question.
  \item If both the option label and option content are available, output the answer in the form of the option label followed by the answer content.
  \item Normalize minor formatting differences, such as extra spaces, redundant punctuation, capitalization, and unnecessary explanatory words.
  \item Preserve important answer details, including numbers, units, variables, signs, option labels, and mathematical expressions.
  \item Do not infer a new answer from the image or question; only canonicalize the given extracted answer.
  \item Do not solve the problem, add reasoning, or change the answer meaning.
\end{itemize}

\medskip
\textbf{Output (strict).} The last line must be a single JSON object with exactly these keys:
\begin{itemize}[leftmargin=*, itemsep=1pt, topsep=1pt]
  \item \texttt{canonical\_answer}, \texttt{answer\_type}, \texttt{recovered\_option\_label}, \texttt{reason}
\end{itemize}
No extra keys. No fences. No text after the JSON.

\medskip
\textbf{JSON schema example.}
\begin{quote}\ttfamily\footnotesize\raggedright\sloppy
\{ "canonical\_answer":"", "answer\_type":"", "recovered\_option\_label":"", "reason":"" \}
\end{quote}

\medskip
\textbf{Example.}
If the question options include ``(D) photosynthesis'', and the extracted answer is ``photosynthesis'', the canonical answer should recover the corresponding option label:
\begin{quote}\ttfamily\footnotesize\raggedright\sloppy
\{ "canonical\_answer":"D. photosynthesis", "answer\_type":"multiple-choice", "recovered\_option\_label":"D", "reason":"The extracted answer matches the content of option D." \}
\end{quote}

\textbf{Image.}
\begin{quote}\ttfamily
<IMAGE\_START>\\
\{image\}\\
<IMAGE\_END>
\end{quote}

\textbf{Refined Question.}
\begin{quote}\ttfamily
<QUESTION\_START>\\
\{question\}\\
<QUESTION\_END>
\end{quote}

\textbf{Extracted Answer.}
\begin{quote}\ttfamily
<ANSWER\_START>\\
\{answer\}\\
<ANSWER\_END>
\end{quote}

\end{tcolorbox}

\refstepcounter{prompt}\label{pmt:domain_classification}
\begin{tcolorbox}[
  colback=gray!5,
  colframe=black!75,
  width=\columnwidth,
  title={Prompt \theprompt: Reasoning Domain Classification},
  breakable,
  boxrule=0.8pt,
  arc=2pt,
  left=6pt,right=6pt,top=6pt,bottom=6pt
]
\small
\raggedright
\sloppy

\textbf{Role.} You are a careful data annotator for multimodal visual reasoning tasks.

\medskip
\textbf{Inputs.} An image, a cleaned question, and a canonical answer.

\medskip
\textbf{Task.} Classify the sample into exactly one reasoning domain according to the primary reasoning skill required to answer the question. Use only the following categories:

\begin{itemize}[leftmargin=*, itemsep=1pt, topsep=1pt]
  \item \texttt{science}: scientific or academic reasoning, including physics, chemistry, biology, geography, astronomy, medicine, scientific diagrams, experimental setups, scientific processes, and knowledge-intensive academic questions.
  
  \item \texttt{math}: mathematical reasoning, including arithmetic, algebra, geometry, probability, equations, formulas, numerical computation, quantitative comparison, and abstract mathematical problem solving.
  
  \item \texttt{chart/table/doc}: reasoning over charts, tables, plots, documents, forms, slides, receipts, reports, text-heavy images, or other structured visual materials that require reading and interpreting visualized or document-like information.
  
  \item \texttt{logic/game/puzzle}: logical reasoning, games, puzzles, rule-based reasoning, pattern reasoning, sequence reasoning, Raven-style problems, visual logic, and tasks that require inferring rules or constraints.
  
  \item \texttt{general}: general visual question answering, including object recognition, scene understanding, attributes, colors, materials, counting without complex computation, common-sense visual understanding, and everyday image questions.
  
  \item \texttt{spatial}: spatial reasoning, including relative positions, directions, distances, object relations, 2D or 3D layouts, orientation, viewpoint understanding, navigation-like reasoning, and geometric relations grounded in the image.
  
  \item \texttt{gui/web}: graphical user interface or web-page understanding, including screenshots, buttons, menus, icons, toolbars, web layouts, interface navigation, and app or browser interaction contexts.
\end{itemize}

\medskip
\textbf{Important Notes.}
\begin{itemize}[leftmargin=*, itemsep=1pt, topsep=1pt]
  \item Choose the domain based on the main skill required to answer the question, not the dataset source or superficial keywords.
  \item If the answer mainly comes from reading a chart, table, plot, document, or text-heavy visual material, prefer \texttt{chart/table/doc}, even if the answer is numeric.
  \item If the task requires scientific knowledge or interpreting a scientific diagram, prefer \texttt{science}; if it mainly requires abstract calculation or formula manipulation, prefer \texttt{math}.
  \item If the task mainly asks about positions, directions, layouts, viewpoints, or spatial relations, prefer \texttt{spatial} over \texttt{general}.
  \item If the image is a software, mobile, desktop, or web interface screenshot, prefer \texttt{gui/web}.
  \item Use \texttt{general} only when the sample does not require specialized scientific, mathematical, logical, spatial, chart/document, or GUI/web reasoning.
  \item Do not solve the problem, modify the question, or change the answer.
\end{itemize}

\medskip
\textbf{Output (strict).} The last line must be a single JSON object with exactly this key:
\begin{itemize}[leftmargin=*, itemsep=1pt, topsep=1pt]
  \item \texttt{domain}
\end{itemize}
The value of \texttt{domain} must be exactly one of the seven categories above. No extra keys. No fences. No text after the JSON.

\medskip
\textbf{JSON schema example.}
\begin{quote}\ttfamily\footnotesize\raggedright\sloppy
\{ "domain":"science" \}
\end{quote}

\textbf{Image.}
\begin{quote}\ttfamily
<IMAGE\_START>\\
\{image\}\\
<IMAGE\_END>
\end{quote}

\textbf{Cleaned Question.}
\begin{quote}\ttfamily
<QUESTION\_START>\\
\{question\}\\
<QUESTION\_END>
\end{quote}

\textbf{Canonical Answer.}
\begin{quote}\ttfamily
<ANSWER\_START>\\
\{answer\}\\
<ANSWER\_END>
\end{quote}

\end{tcolorbox}

\subsection{SFT Trajectory Construction}

\refstepcounter{prompt}\label{pmt:trajectory_distillation}
\begin{tcolorbox}[
  colback=gray!5,
  colframe=black!75,
  width=\columnwidth,
  title={Prompt \theprompt: Reasoning Trajectory Distillation},
  breakable,
  boxrule=0.8pt,
  arc=2pt,
  left=6pt,right=6pt,top=6pt,bottom=6pt
]
\small
\raggedright
\sloppy

\textbf{Image.}
\begin{quote}\ttfamily
\{image\}
\end{quote}

You are a multimodal reasoning assistant. Solve the problem using all relevant information from the image and the question.

\medskip
Your response will be used as a high-quality reasoning trajectory for training another model. Therefore, your reasoning should be correct, clear, concise, and grounded in the image and question.

\medskip
Please follow these requirements:

\begin{enumerate}[leftmargin=*, itemsep=2pt, topsep=2pt]
  \item Carefully understand the task.
  \begin{itemize}[leftmargin=*, itemsep=1pt, topsep=1pt]
    \item Identify what the question asks.
    \item If the question contains multiple sub-questions, answer all of them.
    \item If the question specifies an answer format, such as an option, coordinate, JSON, bbox, point, action format, or LaTeX format such as \texttt{\textbackslash boxed\{\}}, follow it exactly in the final answer.
  \end{itemize}

  \item Use the visual information.
  \begin{itemize}[leftmargin=*, itemsep=1pt, topsep=1pt]
    \item Inspect the image carefully.
    \item Use relevant diagrams, charts, tables, screenshots, UI elements, labels, coordinates, symbols, spatial relations, and visual details.
    \item Do not ignore the image when the question depends on it, or invent unsupported visual details.
  \end{itemize}

  \item Reason clearly and with sufficient detail.
  \begin{itemize}[leftmargin=*, itemsep=1pt, topsep=1pt]
    \item Write the reasoning process before the final answer line.
    \item The reasoning should directly solve the problem and remain concise, logical, and useful.
    \item Avoid unnecessary repetition, irrelevant content, or discussions about the prompt and output format.
    \item For mathematical or scientific problems, include necessary calculations, units, and logical steps.
    \item For multiple-choice questions, compare the options when useful and select the best answer.
    \item For visual grounding, point, bbox, GUI, or action tasks, identify the target carefully and follow the required output schema.
  \end{itemize}

  \item Give the final answer.
  \begin{itemize}[leftmargin=*, itemsep=1pt, topsep=1pt]
    \item After the reasoning process, write the final answer in the required format.
    \item The final answer must be placed inside \texttt{<answer>...</answer>}.
    \item If the problem specifies a required final answer format, such as \texttt{\textbackslash boxed\{\}}, JSON, bbox, \texttt{point\_2d}, option letter, coordinate, or action dictionary, put that format inside \texttt{<answer>...</answer>}.
    \item The content inside \texttt{<answer>...</answer>} should contain only the final answer, such as an option, number, short phrase, coordinate, JSON, bbox, action dictionary, or multiple sub-question answers.
  \end{itemize}
\end{enumerate}

\medskip
\textbf{Problem.}
\begin{quote}\ttfamily
\{question\}
\end{quote}

Your response must follow this order:

\medskip
Write the complete and correct reasoning process first. Then write the final answer line.

\medskip
The last line of your response must match this template exactly:
\begin{quote}\ttfamily\footnotesize\raggedright\sloppy
Therefore, the final answer is <answer>ANSWER</answer>
\end{quote}

Replace ANSWER with the actual solution to the problem.

Do not write anything after the final answer line.

\end{tcolorbox}

\refstepcounter{prompt}\label{pmt:distill_verification}
\begin{tcolorbox}[
  colback=gray!5,
  colframe=black!75,
  width=\columnwidth,
  title={Prompt \theprompt: Distillation Trajectory Verification},
  breakable,
  boxrule=0.8pt,
  arc=2pt,
  left=6pt,right=6pt,top=6pt,bottom=6pt
]
\small
\raggedright
\sloppy

\textbf{System Prompt.}
\begin{quote}\ttfamily\footnotesize\raggedright\sloppy
You are an impartial grading expert. Evaluate whether the candidate response correctly solves the given problem. Final-answer correctness is the primary criterion, while the reasoning trajectory should be considered as a secondary consistency check.
\end{quote}

\medskip
\textbf{Role.}
Given an original question, a standard answer, and a complete candidate response, determine whether the response should be regarded as correct. The candidate response may contain both a reasoning trajectory and a final answer.

\medskip
\textbf{Evaluation Protocol.}

\begin{enumerate}[leftmargin=*, itemsep=2pt, topsep=2pt]

  \item \textbf{Response Validity.}
  \begin{itemize}[leftmargin=*, itemsep=1pt, topsep=1pt]
    \item Reject responses that are incomplete, severely truncated, repetitive, or explicit refusals.
    \item Minor stylistic issues, redundancy, or harmless formatting differences should not affect the judgment.
  \end{itemize}

  \item \textbf{Final-Answer Correctness --- Primary Criterion.}
  \begin{itemize}[leftmargin=*, itemsep=1pt, topsep=1pt]
    \item First determine whether the candidate's final answer is semantically equivalent to the standard answer.
    \item Equivalent mathematical expressions, reasonable numerical precision differences, and semantically equivalent textual answers should be accepted.
    \item For multiple-choice questions, compare the selected option and its corresponding content with the standard answer.
    \item For multi-part questions, all required parts must be correctly answered.
    \item If the final answer is incorrect, classify the response as \texttt{incorrect} regardless of the preceding reasoning.
  \end{itemize}

  \item \textbf{Reasoning Consistency --- Secondary Criterion.}
  \begin{itemize}[leftmargin=*, itemsep=1pt, topsep=1pt]
    \item If the final answer is correct, further examine whether the reasoning trajectory is generally consistent with the question, visual evidence, and final answer.
    \item Check for major logical contradictions, mathematical errors, unsupported visual claims, or factual errors that substantially undermine the validity of the solution.
    \item Minor omissions, shortcuts, imprecise wording, or non-essential intermediate imperfections may be tolerated if they do not affect the validity of the overall solution.
    \item If the trajectory contains a substantive reasoning error that invalidates the solution process, classify the response as \texttt{incorrect} even if the final answer matches the standard answer.
  \end{itemize}

  \item \textbf{Overall Judgment.}
  \begin{itemize}[leftmargin=*, itemsep=1pt, topsep=1pt]
    \item Classify the response as \texttt{correct} when the final answer is correct and the reasoning contains no major inconsistency.
    \item Otherwise, classify the response as \texttt{incorrect}.
  \end{itemize}

\end{enumerate}

\medskip
\textbf{Original Question.}
\begin{quote}\ttfamily\footnotesize\raggedright\sloppy
<Original Question Begin>\\
\{question\}\\
<Original Question End>
\end{quote}

\textbf{Standard Answer.}
\begin{quote}\ttfamily\footnotesize\raggedright\sloppy
<Standard Answer Begin>\\
\{gold\_answer\}\\
<Standard Answer End>
\end{quote}

\textbf{Candidate Response.}
\begin{quote}\ttfamily\footnotesize\raggedright\sloppy
<Candidate Response Begin>\\
\{llm\_response\}\\
<Candidate Response End>
\end{quote}

\medskip
\textbf{Output Format.}
\begin{quote}\ttfamily\footnotesize\raggedright\sloppy
Return exactly one valid JSON object on a single line. Do not output Markdown, code fences, or any additional text.\\

\{"verdict":"correct|incorrect","reasoning\_consistent":true|false,"reason":"brief justification"\}
\end{quote}

\medskip
\textbf{Final Instruction.}
\begin{quote}\ttfamily\footnotesize\raggedright\sloppy
Evaluate the final answer first and use the reasoning trajectory as a secondary consistency check. Return only the required single-line JSON object.
\end{quote}

\end{tcolorbox}

\section{Data Examples}
\label{app:data_examples}

We show representative examples that reflect the diversity and broad task range of MMVistaReason.

\refstepcounter{dataexample}\label{ex:data_science}
\begin{tcolorbox}[
  colback=gray!5,
  colframe=black!75,
  width=\columnwidth,
  title={Example \thedataexample: Science},
  breakable,
  boxrule=0.8pt,
  arc=2pt,
  left=6pt,right=6pt,top=6pt,bottom=6pt
]
\small
\raggedright
\sloppy

\textbf{Image.}
\begin{center}
\includegraphics[width=0.75\columnwidth]{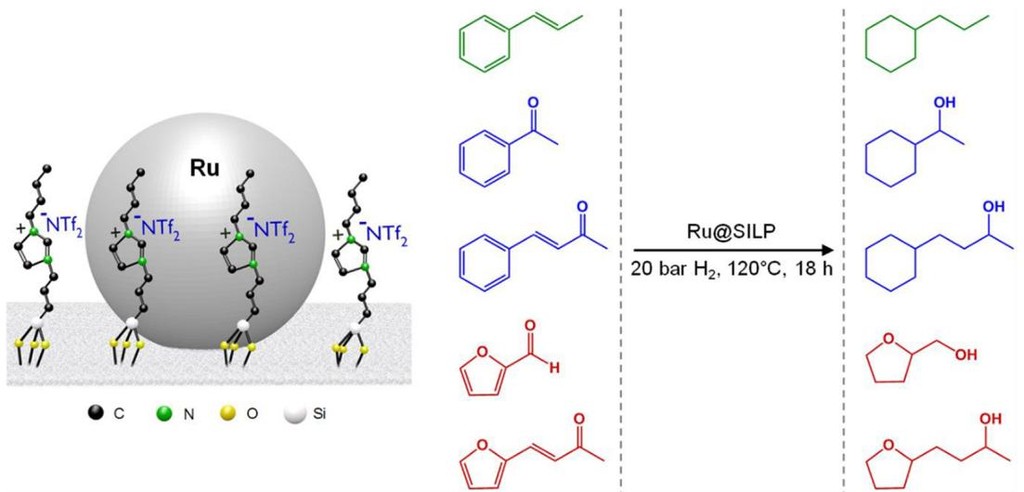}
\end{center}

\textbf{ID.}
\begin{quote}\ttfamily\footnotesize\raggedright\sloppy
scimm\_752608
\end{quote}

\textbf{System Prompt.}
\begin{quote}\ttfamily\footnotesize\raggedright\sloppy
You are a helpful assistant. When answering the question, first provide the reasoning process enclosed in \textless think\textgreater...\textless/think\textgreater{} tags. Then provide the final answer enclosed in \textless answer\textgreater...\textless/answer\textgreater{} tags.
\end{quote}

\textbf{Question.}
\begin{quote}\ttfamily\footnotesize\raggedright\sloppy
<image>\\
What type of product is formed from the hydrogenation of furfuralacetone according to the analysis text?
\end{quote}

\textbf{Answer.}
\begin{quote}\ttfamily\footnotesize\raggedright\sloppy
Saturated alcohol
\end{quote}

\textbf{Metadata.}
\begin{quote}\ttfamily\footnotesize\raggedright\sloppy
source: scimm\_keep\\
domain: science\\
type: close\\
subtype: short\_phrase\\
distill\_model: qwen3.5-27b\\
difficulty: 0.25
\end{quote}

\textbf{Response.}
\begin{quote}\ttfamily\footnotesize\raggedright\sloppy
<think>\\
The image displays a catalytic hydrogenation reaction scheme using Ru@SILP (ruthenium on a silica-immobilized liquid phase catalyst) under 20 bar H$_2$, 120$^\circ$C, for 18 hours. On the left are five substrate molecules; on the right are their corresponding products after hydrogenation.\\

Furfuralacetone is identified as the red molecule in the bottom row of substrates: it has a furan ring attached to an unsaturated ketone chain --- specifically, 4-(furan-2-yl)-3-buten-2-one (also known as furylacrolein or furfuralacetone). Its structure includes a furan ring and an $\alpha,\beta$-unsaturated ketone group.\\

Looking at the product column on the far right, directly aligned with furfuralacetone (bottom red reactant), the product shown is a saturated alcohol: the furan ring remains intact, but the double bond in the side chain is reduced, and the ketone is reduced to a secondary alcohol. The resulting molecule is 4-(furan-2-yl)butan-2-ol.\\

This transformation corresponds to full hydrogenation of both the C=C double bond and the C=O carbonyl group of the enone system, yielding a saturated alcohol while preserving the heteroaromatic furan ring.\\

Thus, according to the visual data in the image, the hydrogenation of furfuralacetone yields a saturated aliphatic alcohol derivative where the enone functionality is fully reduced.\\
</think>\\
Therefore, the final answer is <answer>saturated alcohol</answer>
\end{quote}

\end{tcolorbox}

\refstepcounter{dataexample}\label{ex:data_math}
\begin{tcolorbox}[
  colback=gray!5,
  colframe=black!75,
  width=\columnwidth,
  title={Example \thedataexample: Math},
  breakable,
  boxrule=0.8pt,
  arc=2pt,
  left=6pt,right=6pt,top=6pt,bottom=6pt
]
\small
\raggedright
\sloppy

\textbf{Image.}
\begin{center}
\includegraphics[width=0.45\columnwidth]{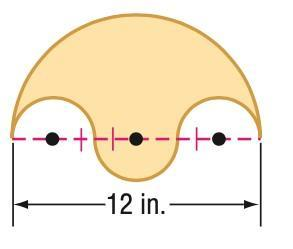}
\end{center}

\textbf{ID.}
\begin{quote}\ttfamily\footnotesize\raggedright\sloppy
mmfr\_113858
\end{quote}

\textbf{System Prompt.}
\begin{quote}\ttfamily\footnotesize\raggedright\sloppy
You are a helpful assistant. When answering the question, first provide the reasoning process enclosed in \textless think\textgreater...\textless/think\textgreater{} tags. Then provide the final answer enclosed in \textless answer\textgreater...\textless/answer\textgreater{} tags.
\end{quote}

\textbf{Question.}
\begin{quote}\ttfamily\footnotesize\raggedright\sloppy
<image>Hint: Please answer the question and provide the correct option letter, e.g., A, B, C, D, at the end.\\
Question: Determine the area of the highlighted section.\\
Choices:\\
(A) 12.6\\
(B) 50.3\\
(C) 100.5\\
(D) 402.1
\end{quote}

\textbf{Answer.}
\begin{quote}\ttfamily\footnotesize\raggedright\sloppy
B
\end{quote}

\textbf{Metadata.}
\begin{quote}\ttfamily\footnotesize\raggedright\sloppy
source: FineVision-geometry3k(mathv360k)\\
domain: math\\
type: close\\
subtype: multiple\_choice\\
distill\_model: qwen3.5-27b\\
difficulty: 1.0
\end{quote}

\textbf{Response.}
\begin{quote}\ttfamily\footnotesize\raggedright\sloppy
<think>\\
To determine the area of the highlighted section, we analyze the shape and its components.\\

The figure consists of:\\
- A large semicircle on top (diameter = 12 in.)\\
- Two small semicircles removed from the bottom left and right (each with diameter = 4 in., since total width is 12 in., and they are symmetrically placed with a middle segment)\\
- One small semicircle added downward in the center (also diameter = 4 in.)\\

Wait --- let's look more carefully at the diagram.\\

Actually, the dashed line shows three points equally spaced along the 12-inch base. The tick marks suggest that each segment between centers is 4 inches. So:\\

- Total width = 12 in.\\
- Divided into 3 equal parts $\rightarrow$ each part is 4 in.\\
- Therefore, the two side ``bites'' are semicircles of diameter 4 in. (radius 2 in.), cut out from the main shape.\\
- The bottom middle protrusion is also a semicircle of diameter 4 in. (radius 2 in.), added to the shape.\\

But wait --- actually, looking again: the overall shape appears to be constructed as follows:\\

Imagine a large rectangle or semicircle? No --- better approach:\\

Notice that the entire shape can be seen as:\\

$\rightarrow$ A large semicircle of diameter 12 in. (area = $\frac{1}{2} \pi r^2 = \frac{1}{2} \pi (6)^2 = 18\pi$)\\

MINUS two small semicircles (on the sides, cut out), each of diameter 4 in. $\rightarrow$ radius 2 in.\\

PLUS one small semicircle at the bottom center (added), also diameter 4 in. $\rightarrow$ radius 2 in.\\

But note: removing two semicircles and adding one net removes one semicircle worth of area.\\

Alternatively, think geometrically:\\

The shape is equivalent to:\\

Large semicircle (d=12) minus one small semicircle (d=4).\\

Why? Because you remove two small semicircles from the top edge (but they're below the diameter?), no --- let's reinterpret.\\

Actually, standard interpretation for such figures (common in geometry problems):\\

This shape is formed by:\\

- Taking a large semicircle of diameter 12 in. (so radius 6 in.)\\
- Subtracting two smaller semicircles (each diameter 4 in., so radius 2 in.) from the flat side --- but those would be indents.\\
- And adding a semicircle of diameter 4 in. hanging down from the center.\\

But visually, the net effect is:\\

Area = Area of large semicircle - Area of two small semicircles + Area of one small semicircle\\
= Area of large semicircle - Area of one small semicircle\\

Because -2 +1 = -1.\\

So:\\

Large semicircle: diameter 12 $\rightarrow$ radius 6 $\rightarrow$ area = $(1/2) * \pi * 6^2 = 18\pi$\\

Small semicircle: diameter 4 $\rightarrow$ radius 2 $\rightarrow$ area = $(1/2) * \pi * 2^2 = 2\pi$\\

Thus, shaded area = $18\pi - 2\pi = 16\pi$\\

Now compute numerically: $\pi \approx 3.1416 \rightarrow 16 * 3.1416 \approx 50.2656 \approx 50.3$\\

Therefore, the area is approximately 50.3 square inches.\\

Matching choice: (B) 50.3\\
</think>\\
Therefore, the final answer is <answer>B</answer>
\end{quote}

\end{tcolorbox}

\refstepcounter{dataexample}\label{ex:data_logic}
\begin{tcolorbox}[
  colback=gray!5,
  colframe=black!75,
  width=\columnwidth,
  title={Example \thedataexample: Logic/Game/Puzzle},
  breakable,
  boxrule=0.8pt,
  arc=2pt,
  left=6pt,right=6pt,top=6pt,bottom=6pt
]
\small
\raggedright
\sloppy

\textbf{Image.}
\begin{center}
\includegraphics[width=0.45\columnwidth]{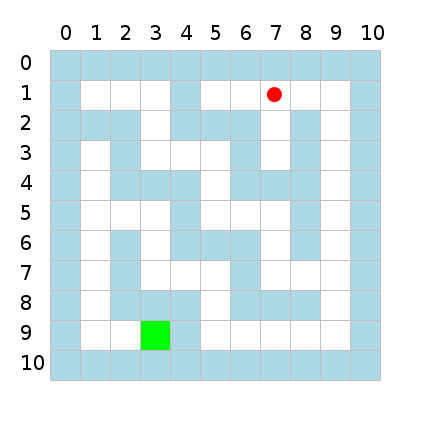}
\end{center}

\textbf{ID.}
\begin{quote}\ttfamily\footnotesize\raggedright\sloppy
mmfr\_304654
\end{quote}

\textbf{System Prompt.}
\begin{quote}\ttfamily\footnotesize\raggedright\sloppy
You are a helpful assistant. When answering the question, first provide the reasoning process enclosed in \textless think\textgreater...\textless/think\textgreater{} tags. Then provide the final answer enclosed in \textless answer\textgreater...\textless/answer\textgreater{} tags.
\end{quote}

\textbf{Question.}
\begin{quote}\ttfamily\footnotesize\raggedright\sloppy
<image>\\
**Rules:**\\
1. This is a maze mini-game. The player needs to navigate around obstacles to reach the destination and achieve victory.\\
2. The red circle represents the player, the green block is the goal and the blue blocks are obstacles.\\
3. The player can only move within the white blocks.\\
4. The coordinates are given in the format (row, col), where row represents the vertical position and col represents the horizontal position.\\

**Question:** Which directions are available to move now?\\

**Options:**\\
A. up\\
B. down\\
C. up, left\\
D. up, right\\
E. up, down, left\\
F. up, down, right\\
G. down, left, right\\
H. up, down, left, right
\end{quote}

\textbf{Answer.}
\begin{quote}\ttfamily\footnotesize\raggedright\sloppy
G
\end{quote}

\textbf{Metadata.}
\begin{quote}\ttfamily\footnotesize\raggedright\sloppy
source: GameQA-140K\\
domain: logic/game/puzzle\\
type: close\\
subtype: multiple\_choice\\
distill\_model: qwen3.5-27b\\
difficulty: 0.75
\end{quote}

\textbf{Response.}
\begin{quote}\ttfamily\footnotesize\raggedright\sloppy
<think>\\
To determine which directions are available to move from the current position of the player (red circle), we need to:\\

1. Identify the coordinates of the red circle.\\
- From the image, the red circle is at row 1, column 7, i.e., (1, 7).\\

2. Check adjacent cells (up, down, left, right) to see if they are white (walkable) or blue (obstacle).\\
- Up: (0, 7). This cell is within the grid and appears white in the maze path? Let us check visually. Looking at row 0, columns 0--10, the top row is entirely light blue, meaning it is an obstacle. So up from (1,7) goes to (0,7), which is blue and not walkable.\\
- Down: (2, 7). In the image, this cell is white and therefore walkable.\\
- Left: (1, 6). This cell is white and therefore walkable.\\
- Right: (1, 8). This cell is also white and therefore walkable.\\

Wait, let me double-check with the actual maze structure shown.\\

Looking at the maze layout more carefully:\\
- At (1,7): red dot.\\
- Above it (0,7): part of the outer border. The entire row 0 is shaded light blue, so that is an obstacle and cannot be used.\\
- Below it (2,7): white square, so the player can go down.\\
- To the left (1,6): white square, so the player can go left.\\
- To the right (1,8): white square, so the player can go right.\\

So available moves are down, left, and right.\\

But wait, option G is ``down, left, right'', which matches.\\

However, let me re-express based on visual inspection:\\

In row 1:\\
- Columns: 0=blue, 1=white, 2=white, 3=blue, 4=white, 5=white, 6=white, 7=red (player), 8=white, 9=white, 10=blue.\\

So from (1,7):\\
- Left: (1,6) = white, so OK.\\
- Right: (1,8) = white, so OK.\\
- Down: (2,7) = white, so OK.\\
- Up: (0,7) = row 0, all blue, so not walkable.\\

Thus, available directions are left, right, and down.\\

That corresponds to option G.\\
</think>\\
Therefore, the final answer is <answer>G</answer>
\end{quote}

\end{tcolorbox}

\refstepcounter{dataexample}\label{ex:data_chart}
\begin{tcolorbox}[
  colback=gray!5,
  colframe=black!75,
  width=\columnwidth,
  title={Example \thedataexample: Chart/Table/Doc},
  breakable,
  boxrule=0.8pt,
  arc=2pt,
  left=6pt,right=6pt,top=6pt,bottom=6pt
]
\small
\raggedright
\sloppy

\textbf{Image.}
\begin{center}
\includegraphics[width=0.75\columnwidth]{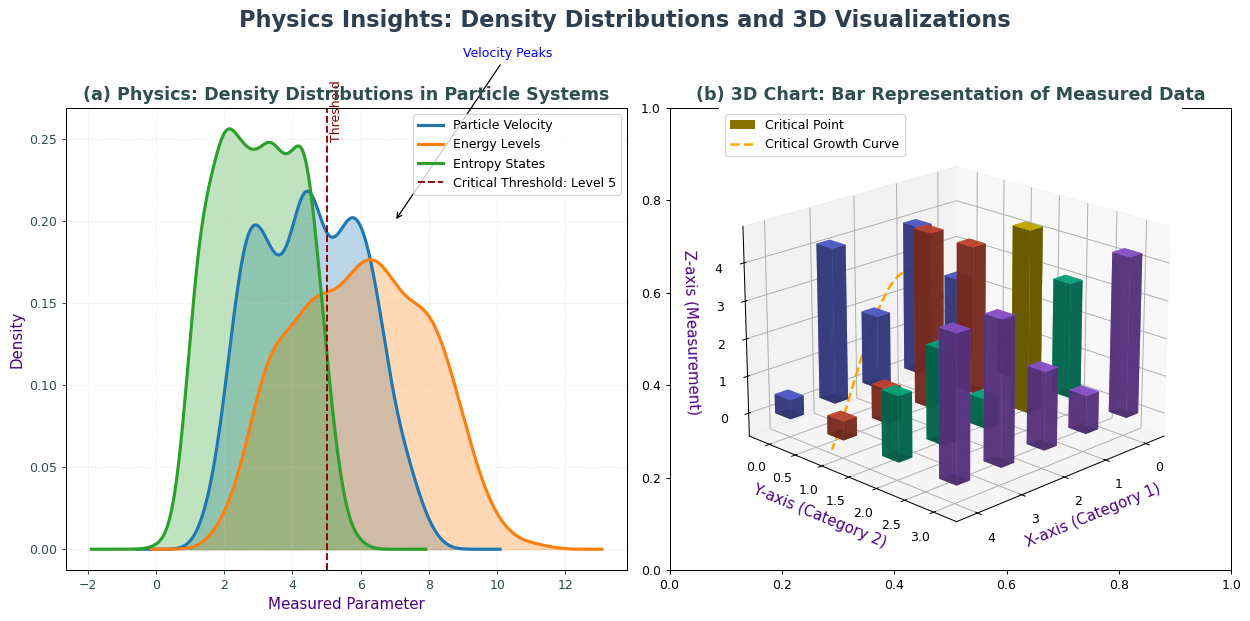}
\end{center}

\textbf{ID.}
\begin{quote}\ttfamily\footnotesize\raggedright\sloppy
vero\_129302
\end{quote}

\textbf{System Prompt.}
\begin{quote}\ttfamily\footnotesize\raggedright\sloppy
You are a helpful assistant. When answering the question, first provide the reasoning process enclosed in \textless think\textgreater...\textless/think\textgreater{} tags. Then provide the final answer enclosed in \textless answer\textgreater...\textless/answer\textgreater{} tags.
\end{quote}

\textbf{Question.}
\begin{quote}\ttfamily\footnotesize\raggedright\sloppy
<image>\\
What is the x-coordinate of the bar with the maximum z-height in subplot (b)?
\end{quote}

\textbf{Answer.}
\begin{quote}\ttfamily\footnotesize\raggedright\sloppy
3
\end{quote}

\textbf{Metadata.}
\begin{quote}\ttfamily\footnotesize\raggedright\sloppy
source: vero\_chart\_ocr-ecd\_vqa\\
domain: chart/table/doc\\
type: close\\
subtype: numeric\\
distill\_model: qwen3.5-27b\\
difficulty: 0.25
\end{quote}

\textbf{Response.}
\begin{quote}\ttfamily\footnotesize\raggedright\sloppy
<think>\\
To solve this problem, we need to examine subplot (b), which is a 3D bar chart titled ``3D Chart: Bar Representation of Measured Data''.\\

The question asks for the **x-coordinate** of the bar with the **maximum z-height**.\\

Step 1: Understand the axes.\\
- The X-axis is labeled ``X-axis (Category 1)'' and ranges from 0 to 4 (approximately).\\
- The Y-axis is labeled ``Y-axis (Category 2)'' and ranges from 0.0 to 3.0.\\
- The Z-axis is labeled ``Z-axis (Measurement)'' and shows the height of each bar.\\

Step 2: Identify the tallest bar visually.\\
Looking at the 3D bar chart in subplot (b), the tallest bar (highest z-value) is the yellow-green bar located near the center-right of the plot. Its z-height appears to be around 5 or slightly above, clearly higher than any other bar.\\

Step 3: Determine its x-coordinate.\\
We project that bar down onto the X-axis. Observing the position along the X-axis (``Category 1''), this bar aligns with approximately **x = 3.0**.\\

Let's verify by checking nearby bars:\\
- At x=2, there are several bars but none reach as high as the one at x=3.\\
- At x=4, the purple bar is tall but not as tall as the yellow-green one at x=3.\\
- The yellow-green bar is positioned directly above x=3 on the X-axis.\\

Thus, the bar with maximum z-height has an x-coordinate of 3.\\
</think>\\
Therefore, the final answer is <answer>3</answer>
\end{quote}

\end{tcolorbox}

\refstepcounter{dataexample}\label{ex:data_general}
\begin{tcolorbox}[
  colback=gray!5,
  colframe=black!75,
  width=\columnwidth,
  title={Example \thedataexample: General},
  breakable,
  boxrule=0.8pt,
  arc=2pt,
  left=6pt,right=6pt,top=6pt,bottom=6pt
]
\small
\raggedright
\sloppy

\textbf{Image.}
\begin{center}
\includegraphics[width=0.4\columnwidth]{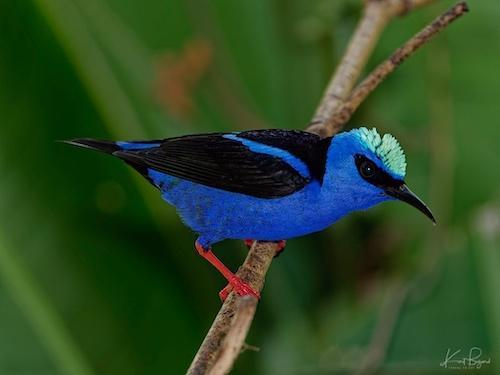}
\end{center}

\textbf{ID.}
\begin{quote}\ttfamily\footnotesize\raggedright\sloppy
vero\_19012
\end{quote}

\textbf{System Prompt.}
\begin{quote}\ttfamily\footnotesize\raggedright\sloppy
You are a helpful assistant. When answering the question, first provide the reasoning process enclosed in \textless think\textgreater...\textless/think\textgreater{} tags. Then provide the final answer enclosed in \textless answer\textgreater...\textless/answer\textgreater{} tags.
\end{quote}

\textbf{Question.}
\begin{quote}\ttfamily\footnotesize\raggedright\sloppy
<image>\\
What kind of bird is in the photo?
\end{quote}

\textbf{Answer.}
\begin{quote}\ttfamily\footnotesize\raggedright\sloppy
Red-legged Honeycreeper (Cyanerpes cyaneus)
\end{quote}

\textbf{Metadata.}
\begin{quote}\ttfamily\footnotesize\raggedright\sloppy
source: vero\_captioning\_IF-pixmo\_ask\_model\_anything\\
domain: general\\
type: close\\
subtype: entity\_name\\
distill\_model: qwen3.5-27b\\
difficulty: 0.5
\end{quote}

\textbf{Response.}
\begin{quote}\ttfamily\footnotesize\raggedright\sloppy
<think>\\
To identify the bird in the photo, we analyze its key visual features:\\

1. **Coloration**: The bird has a striking electric blue body, which is very distinctive.\\
2. **Head crest**: It displays a bright turquoise-green crest on top of its head, standing erect --- a hallmark feature.\\
3. **Wings and back**: The wings are black with a vivid blue stripe running along the edge (coverts), contrasting sharply with the blue body.\\
4. **Legs**: The legs are bright red or orange-red, another notable trait.\\
5. **Beak**: Short, pointed, dark beak suited for nectar feeding.\\
6. **Tail**: Relatively short tail, also mostly black with some blue at the base.\\

These characteristics --- especially the combination of brilliant blue plumage, greenish-blue crest, black wings with blue edges, and red legs --- are diagnostic of the **Red-legged Honeycreeper** (*Cyanerpes cyaneus*).\\

This species is found from Mexico through Central America to northern South America, often in tropical forests and gardens. Males are brightly colored as seen here; females are more olive-green with less vibrant crests.\\

The image clearly shows an adult male due to the intense coloration and prominent crest.\\
</think>\\
Therefore, the final answer is <answer>Red-legged Honeycreeper</answer>
\end{quote}

\end{tcolorbox}

\refstepcounter{dataexample}\label{ex:data_spatial}
\begin{tcolorbox}[
  colback=gray!5,
  colframe=black!75,
  width=\columnwidth,
  title={Example \thedataexample: Spatial},
  breakable,
  boxrule=0.8pt,
  arc=2pt,
  left=6pt,right=6pt,top=6pt,bottom=6pt
]
\small
\raggedright
\sloppy

\textbf{Image.}
\begin{center}
\includegraphics[width=0.5\columnwidth]{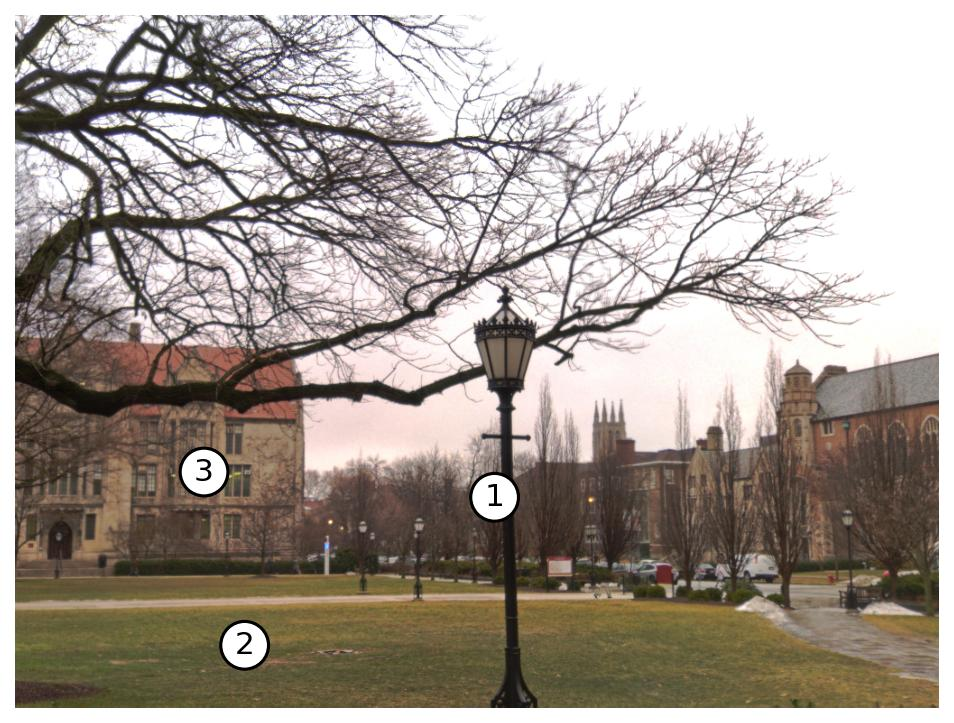}
\end{center}

\textbf{ID.}
\begin{quote}\ttfamily\footnotesize\raggedright\sloppy
vero\_400822
\end{quote}

\textbf{System Prompt.}
\begin{quote}\ttfamily\footnotesize\raggedright\sloppy
You are a helpful assistant. When answering the question, first provide the reasoning process enclosed in \textless think\textgreater...\textless/think\textgreater{} tags. Then provide the final answer enclosed in \textless answer\textgreater...\textless/answer\textgreater{} tags.
\end{quote}

\textbf{Question.}
\begin{quote}\ttfamily\footnotesize\raggedright\sloppy
<image>\\
The original image has three regions marked as 1, 2, and 3. Consider the content, positional relationships, depths of the three regions and other cues, and sort the depths of the three regions from smallest to largest from the camera, where the format of the answer is ``Smallest, Middle, Largest'' (e.g., ``1, 2, 3'').
\end{quote}

\textbf{Answer.}
\begin{quote}\ttfamily\footnotesize\raggedright\sloppy
2, 1, 3
\end{quote}

\textbf{Metadata.}
\begin{quote}\ttfamily\footnotesize\raggedright\sloppy
source: vero\_spatial\_action-spatial\_ssrl\\
domain: spatial\\
type: close\\
subtype: position\\
distill\_model: qwen3.5-122b\\
difficulty: 0.5
\end{quote}

\textbf{Response.}
\begin{quote}\ttfamily\footnotesize\raggedright\sloppy
<think>\\
To determine the relative depths of regions 1, 2, and 3 from the camera:\\

- Region 1 is a lamppost located in the foreground, clearly closer to the viewer than the buildings or distant trees. It stands on the grassy area but is positioned such that it partially overlaps with background elements, indicating it is nearer.\\

- Region 2 is a patch of lawn in the immediate foreground, extending toward the bottom of the frame. Since it lies directly beneath the lamppost (region 1) and extends into the scene without obstruction, it is actually closer to the camera than region 1, because the base of the lamppost emerges from this lawn, meaning the lawn starts at the very front edge of the image.\\

Wait, let us re-evaluate based on visual depth cues:\\

Actually, looking again:\\

Region 2 is labeled on the grass in the lower-left portion of the image. This is part of the open lawn that stretches from the bottom edge of the photo toward the middle ground. The lamppost (region 1) is standing on this lawn, slightly further back. You can see the base of the lamppost embedded in the grass near region 2. So region 2 is closer to the camera than region 1.\\

Region 3 is on the facade of a large stone building in the mid-to-background. This building is clearly behind both the lawn and the lamppost. You can tell by perspective: the building appears smaller due to distance, and other objects like trees and pathways lie between the camera and the building.\\

Additionally, the tree branches overhead span across the top of the image. They are likely in the extreme foreground, even closer than region 2, but since we are only comparing regions 1, 2, and 3, we ignore them for ranking.\\

So ordering by depth, where smallest means closest:\\

- Region 2: closest, because it is the foreground lawn right at the bottom of the frame.\\
- Region 1: middle, because the lamppost stands on the lawn, so it is farther than region 2 but still in the foreground or midground.\\
- Region 3: farthest, because it is on the distant building.\\

Thus, sorted from smallest (closest) to largest (farthest) depth: 2, 1, 3.\\
</think>\\
Therefore, the final answer is <answer>2, 1, 3</answer>
\end{quote}

\end{tcolorbox}

\refstepcounter{dataexample}\label{ex:data_guiweb}
\begin{tcolorbox}[
  colback=gray!5,
  colframe=black!75,
  width=\columnwidth,
  title={Example \thedataexample: GUI/Web},
  breakable,
  boxrule=0.8pt,
  arc=2pt,
  left=6pt,right=6pt,top=6pt,bottom=6pt
]
\small
\raggedright
\sloppy

\textbf{Image.}
\begin{center}
\includegraphics[width=0.5\columnwidth]{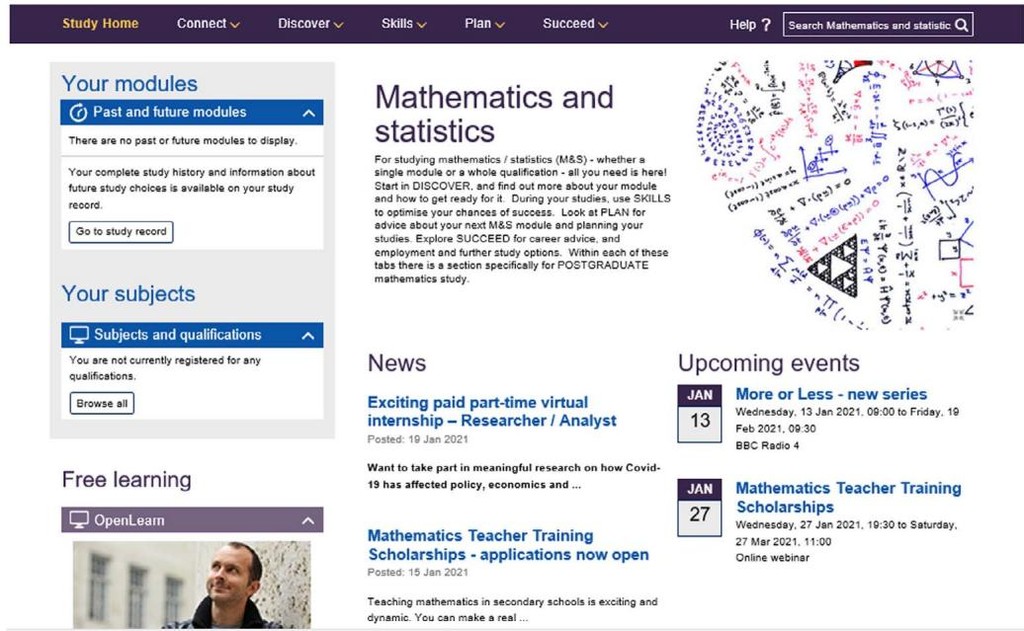}
\end{center}

\textbf{ID.}
\begin{quote}\ttfamily\footnotesize\raggedright\sloppy
scimm\_847127
\end{quote}

\textbf{System Prompt.}
\begin{quote}\ttfamily\footnotesize\raggedright\sloppy
You are a helpful assistant. When answering the question, first provide the reasoning process enclosed in \textless think\textgreater...\textless/think\textgreater{} tags. Then provide the final answer enclosed in \textless answer\textgreater...\textless/answer\textgreater{} tags.
\end{quote}

\textbf{Question.}
\begin{quote}\ttfamily\footnotesize\raggedright\sloppy
<image>\\
How many customized sections follow the generic template on the M\&S study site?
\end{quote}

\textbf{Answer.}
\begin{quote}\ttfamily\footnotesize\raggedright\sloppy
5
\end{quote}

\textbf{Metadata.}
\begin{quote}\ttfamily\footnotesize\raggedright\sloppy
source: scimm\_keep\\
domain: gui/web\\
type: close\\
subtype: counting\\
distill\_model: qwen3.5-9b\\
difficulty: 0.25
\end{quote}

\textbf{Response.}
\begin{quote}\ttfamily\footnotesize\raggedright\sloppy
<think>\\
To determine how many customized sections follow the generic template on the M\&S (Mathematics and Statistics) study site, we analyze the structure of the page shown in the image.\\

The page is organized into several distinct functional areas or ``sections'' that appear after the top navigation bar, including ``Study Home'' and ``Connect'', which can be considered the generic template. These sections include:\\

1. **Your modules** -- Contains information about past/future modules and a link to study record.\\
2. **Your subjects** -- Shows subject/qualification registration status and a browse button.\\
3. **Free learning** -- Features OpenLearn content with an accompanying image.\\
4. **News** -- Displays posted updates, such as internship announcement and scholarship information.\\
5. **Upcoming events** -- Lists scheduled events with dates and descriptions.\\

Each of these five sections has its own content, layout, and purpose, suggesting they are customized for user interaction and information delivery on this specific M\&S study portal.\\

No other distinct section types beyond these five are visible on the main content area below the header.\\

Thus, there are **five** customized sections following the generic template.\\
</think>\\
Therefore, the final answer is <answer>5</answer>
\end{quote}

\end{tcolorbox}

\end{document}